\documentclass[sigconf,nonacm]{acmart}
\AtBeginDocument{
  }

\usepackage{bm}

\acmDOI{XXXXXXX.XXXXXXX}
\acmISBN{978-1-4503-XXXX-X/2018/06}

\newcommand{\Kxy}{K^{xy}}
\newcommand{\gpt}{GPT-4o}
\newcommand{\gem}{Gem-2f}

\begin{document}

\title{LLMs Reproduce Human Purchase Intent via Semantic Similarity Elicitation of Likert Ratings}

\author{Benjamin F. Maier}
\email{ben.maier@pymc-labs.com}
\orcid{0000-0001-7414-8823}
\affiliation{
  \institution{\href{https://www.pymc-labs.com}{PyMC Labs}}
  \country{\vspace{-3mm}\ }
}
\authornote{Corresponding authors.}
\author{Ulf Aslak}
\email{ulf.aslak@pymc-labs.com}
\orcid{0000-0003-4704-3609}
\affiliation{
  \institution{\href{https://www.pymc-labs.com}{PyMC Labs}}
  \country{\vspace{-3mm}\ }
}
\author{Luca~Fiaschi}
\email{luca.fiaschi@pymc-labs.com}
\affiliation{
  \institution{\href{https://www.pymc-labs.com}{PyMC Labs}}
  \country{\vspace{-3mm}\ }
}
\author{Nina~Rismal}
\email{nina.rismal@pymc-labs.com}
\affiliation{
  \institution{\href{https://www.pymc-labs.com}{PyMC Labs}}
  \country{\vspace{-3mm}\ }
}
\author{Kemble~Fletcher}
\email{kemble.fletcher@pymc-labs.com}
\affiliation{
  \institution{\href{https://www.pymc-labs.com}{PyMC Labs}}
  \country{\vspace{-3mm}\ }
}
\author{Christian~C.~Luhmann}
\email{christian.luhmann@pymc-labs.com}
\orcid{0000-0002-9773-1672}
\affiliation{
  \institution{\href{https://www.pymc-labs.com}{PyMC Labs}}
  \country{\vspace{-3mm}\ }
}
\author{Robbie~Dow}
\email{robbie_dow@colpal.com}
\orcid{0009-0009-2484-2230}
\affiliation{
  \institution{Colgate-Palmolive Company}
  \city{New York}
  \state{NY}
  \country{USA}
}
\author{Kli~Pappas}
\email{kli_pappas@colpal.com}
\affiliation{
  \institution{Colgate-Palmolive Company}
  \city{New York}
  \state{NY}
  \country{USA}
}
\authornotemark[1]
\author{Thomas~V.~Wiecki}
\email{thomas.wiecki@pymc-labs.com}
\orcid{0009-0000-6015-101X}
\affiliation{
  \institution{\href{https://www.pymc-labs.com}{PyMC Labs}}
  \country{\ }
}

\renewcommand{\shortauthors}{Maier \textit{et al.}}

\begin{abstract}
Consumer research costs companies billions annually yet suffers from panel biases and limited scale. Large language models (LLMs) offer an alternative by simulating synthetic consumers, but produce unrealistic response distributions when asked directly for numerical ratings. We present semantic similarity rating (SSR), a method that elicits textual responses from LLMs and maps these to Likert distributions using embedding similarity to reference statements. Testing on an extensive dataset comprising 57 personal care product surveys conducted by a leading corporation in that market (9,300 human responses), SSR achieves 90\% of human test--retest reliability while maintaining realistic response distributions (KS similarity $>0.85$). Additionally, these synthetic respondents provide rich qualitative feedback explaining their ratings. This framework enables scalable consumer research simulations while preserving traditional survey metrics and interpretability.
\end{abstract}



\received{\today}
\maketitle

\section{Introduction}
Established consumer research plays a central role in guiding corporations' product development decisions \cite{keller2015marketing, ulrich2015product, esomar2024}, costing them billions of U.S.~dollars globally every year \cite{esomar2024}.
Before investing in costly production and launch activities, companies routinely evaluate product concepts by surveying representative consumer panels.
The most consequential question in such studies typically concerns purchase intent (PI), i.e., the likelihood that a respondent would buy the product if available \cite{clancy2001market, silk1976pretest, jamieson1989purchase}.
Standard practice is to elicit purchase intent on a Likert scale, usually ranging from 1 (e.g., ``definitely not'') to 5 (e.g., ``definitely yes'') \cite{likert1932technique}.
While widely used, this method faces well-known limitations: responses may be distorted by satisficing, acquiescence, and positivity biases, among other factors \cite{krosnick1991satisficing, krosnick1999survey}.
Thus, traditional consumer panels often provide noisy measurements of demand, despite the considerable resources invested.

Recent advances in LLMs raise the possibility of augmenting or partially replacing human survey panels with synthetic consumers.
By conditioning LLMs on demographic or attitudinal personas and exposing them to the same survey instruments, researchers have begun exploring whether such synthetic samples can recover human-like patterns of response.
This line of work has expanded rapidly across disciplines, including market research, political science, psychology, and consumer behavior \cite{li2024validity,brand2024llmmarketresearch,argyle2023outofone,bisbee2024perils,kaiser2025simulating,eichstaedt2023social}.
Taken together, this literature establishes the prominence of LLM-based synthetic samples while also underscoring challenges regarding their reliability.

One recurring challenge is the direct elicitation of Likert-scale responses.
When asked to provide numerical ratings, LLMs tend to produce distributions that are overly narrow, systematically skewed, or otherwise inconsistent with human survey data \cite{bisbee2024perils,kaiser2025simulating,eichstaedt2023social}.
This raises the question of whether such shortcomings reflect fundamental limits of LLMs as survey respondents, or simply the elicitation methods used.

In this paper, we argue for the latter.
We show that the problem lies not in the use of LLMs themselves, but in directly requesting Likert-scale outputs.
Instead, we propose to use textual elicitation followed by semantic-similarity rating (SSR): LLMs generate free-text statements of purchase intent, which are then projected onto a 5-point (5pt) Likert scale by computing the cosine similarity of embeddings with those of predefined anchor statements.
This approach draws on established methods in NLP (semantic similarity mapping) \cite{yin2019benchmarking} and survey methodology (anchoring vignettes) \cite{king2004enhancing}, but to our knowledge has not been applied in the context of LLMs as survey respondents.

Using 57 consumer research surveys on personal care product concepts conducted by a leading corporation in that market, each with 150–400 human participants, we demonstrate that SSR closely replicates real survey outcomes.
Specifically, it recovers both (1) the panel-level response distributions and (2) the relative ranking of product concepts by mean purchase intent. To quantify the latter, we introduce \emph{correlation attainment} herein which is inspired by human test--retest reliability experiments and measures correlation between synthetic and real data in terms of its maximum achievable value.
Moreover, we demonstrate that these results are only achieved when LLMs are prompted to consider demographic attributes of a person they are being asked to impersonate. We find that the LLMs' response behavior with regard to age and income level, in particular, mirrors actual humans' response behavior relatively well. As a byproduct of SSR, feedback on product concepts can be leveraged for qualitative analyses and further concept development.

These results suggest that, when paired with appropriate elicitation methods, LLMs can serve as valid synthetic consumers for concept testing.

\section{Related Works}

Several studies using LLMs as synthetic survey respondents rely on direct numeric elicitation. For example, models are asked to fill in Likert scales \cite{eichstaedt2023social} or provide ``feeling thermometer'' scores \cite{bisbee2024perils}. Others adapt this approach to discrete-choice tasks, such as conjoint-style willingness-to-pay estimation \cite{brand2024llmmarketresearch} or behavioral games \cite{aher2023turing}. A consistent limitation is that response distributions are too narrow: models regress to ``typical'' answers, showing far less variance than human data and producing unrealistically confident estimates \cite{bisbee2024perils,aher2023turing}.

A smaller line of work explores textual responses that are subsequently mapped onto numbers. One study uses elicitation of open completions (``[Brand] is similar to...'') and converts those into similarity scores by counting elicited brand completions \cite{li2024validity}. Another trains ``Doppelg\"anger LLMs'' on individual utterances, generating free-text survey answers that are then aligned with structured categories \cite{cho2025}. While such pipelines acknowledge the ambiguity of open-ended responses, they ultimately reduce them back to single numbers.

Another focus of some studies is demographic conditioning, where prompts embed socio-demographic backstories. One study shows that this improves the alignment of synthetic subgroup responses with human benchmarks \cite{argyle2023outofone}, and another demonstrates similar effects with political personas \cite{bisbee2024perils}. Conditioning increases validity but does not overcome the fundamental issue of narrow distributions.

Some work uses fine-tuning with survey data to make LLMs more human-like \cite{argyle2023outofone, brand2024llmmarketresearch,cho2025}. But a large share of the literature, including refs.\ \cite{bisbee2024perils, li2024validity, jansen2023employing, aher2023turing}, stays with zero-shot elicitation or prompt engineering.

\begin{figure*}
    \centering
    \includegraphics[width=\textwidth]{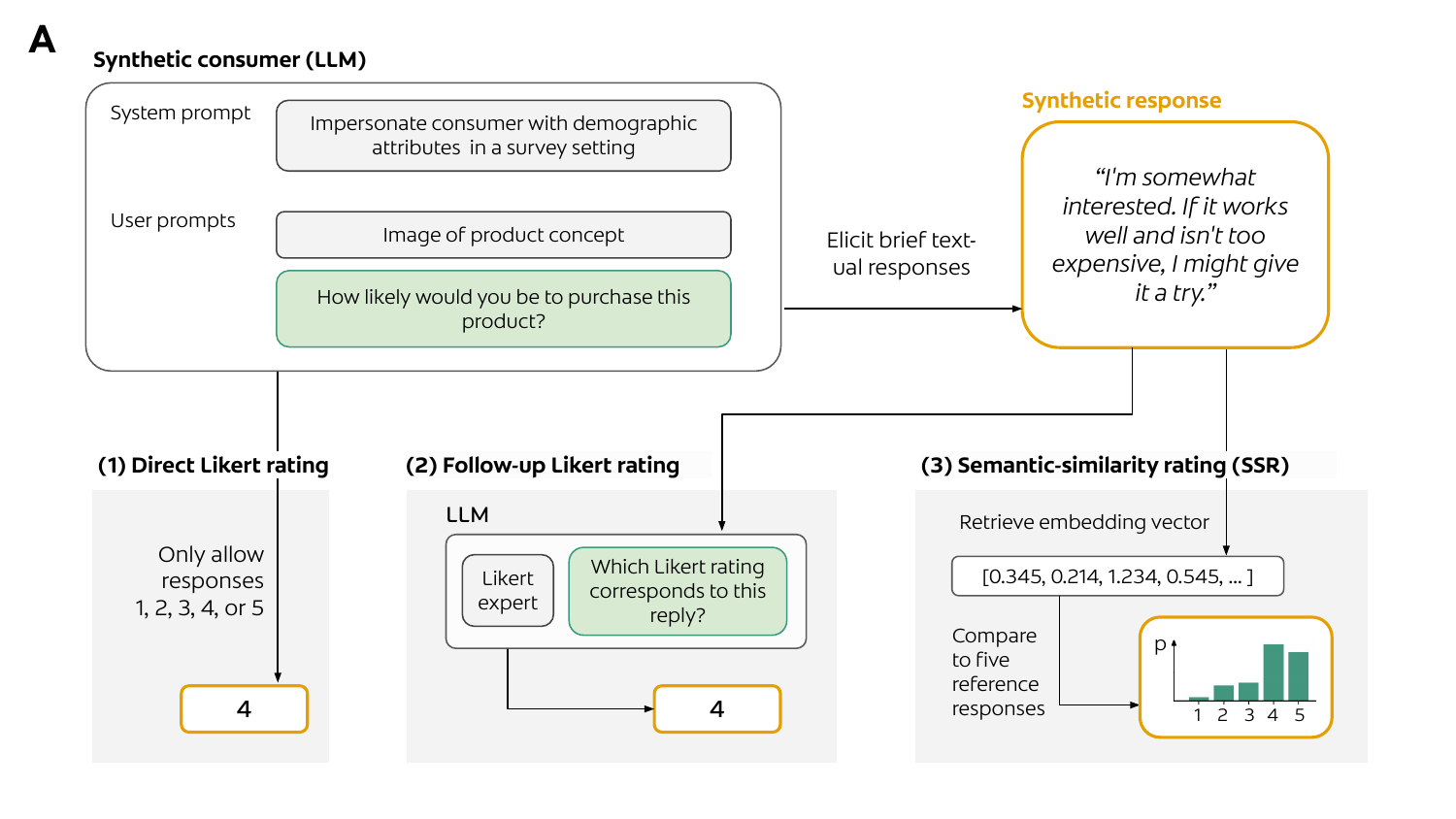} 

    \includegraphics[width=\textwidth]{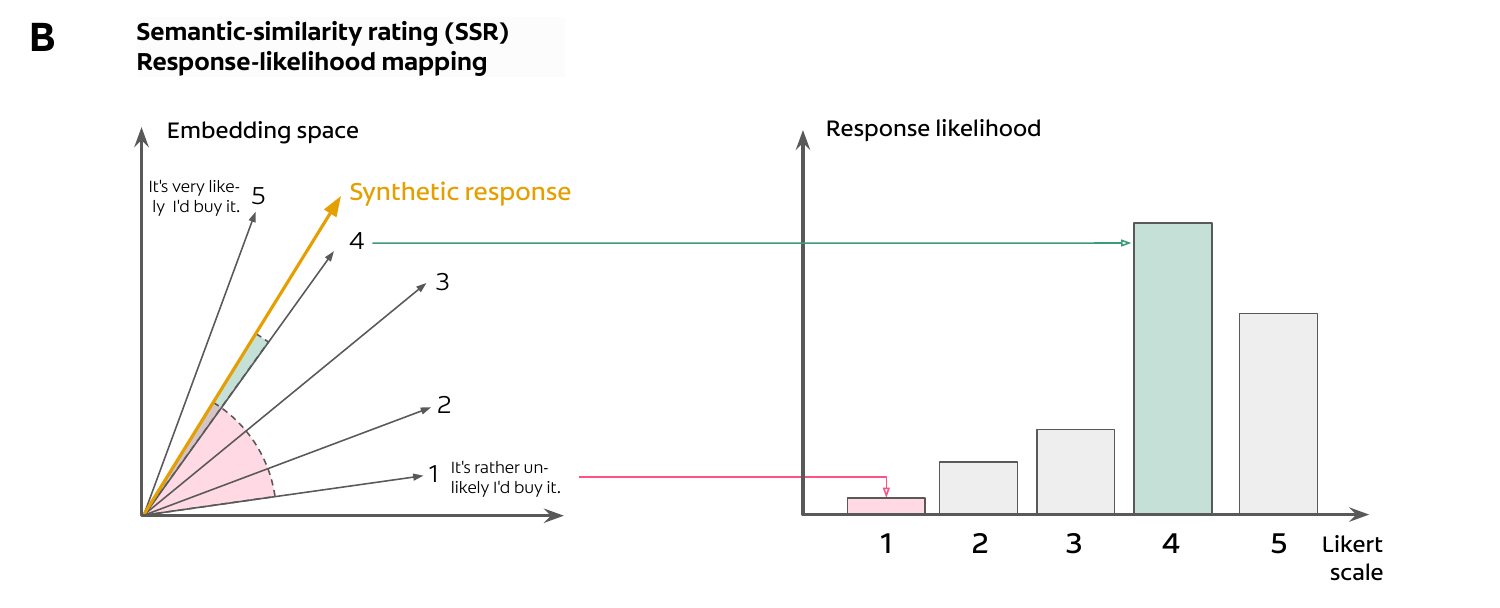} 

    \caption{Different response generation procedures and SSR response-likelihood mapping. \textbf{(A)} A synthetic consumer is constructed by instructing an LLM to impersonate a consumer with certain demographic properties and show them a product concept as an image containing a description and possibly concept art (see App.~\ref{sec:product-concept-examples}). The synthetic consumer is then asked about their purchase intent. (1) In the direct Likert-rating approach, the LLM's response is restricted to one of 1, 2, 3, 4, or 5. (2) Alternatively, we let the LLM write a brief textual response about their PI. Subsequently, we prompt the same model to be a Likert-rating ``expert'' and map the textual response to an integer between 1 and 5. (3) Because textual responses can result in varied ratings on the 5pt Likert scale, we introduce the semantic similarity rating method. We retrieve the embedding vector for the textual response from a corresponding model, compare it to five reference response embedding vectors and construct a response distribution on the Likert scale.
    \textbf{(B)} In an embedding space, the synthetic response will have a certain angular distance to any other statement. We construct a reference set of five rating responses, each corresponding to an integer on the Likert scale. Then, the response likelihood of any integer is set to be proportional to the cosine similarity between the synthetic response vector and the corresponding reference response vector.
    }
    \label{fig:response-generation-and-esr}
\end{figure*}

\section{Methods}
This section provides a brief overview of the methods employed in this paper. See App.~\ref{app:methods} for a detailed explanation.

\subsection{Data}
We analyze 57 consumer research surveys on personal care product concepts conducted by a leading corporation in that market using a digital consumer research platform (see App.~\ref{app:data}). Each survey involved 150--400 unique U.S. participants ($N=9{,}300$ in total), with core demographic markers such as age, gender, and location available for most respondents, and income and ethnicity for fewer. Surveys asked participants to evaluate a concept and rate their purchase intent on a 5pt Likert scale. Mean purchase intent is skewed towards positive values and narrowly distributed with mean $4.0$ and standard deviation $0.1$ across all surveys.

\subsection{Definitions}
Following the definitions in App.~\ref{app:definitions}, each survey $s$ is associated with a product concept and a set of consumers $c \in \mathcal{C}_s$, who each provide a Likert rating $r_c \in \{1,\dots,5\}$ marking their purchase intent. Per survey, these form empirical response distributions and mean purchase intent 
$\mathrm{PI}_s$.
We define synthetic consumers $c'$ as LLMs impersonating human respondents $c$ given their demographic attributes. Unlike real consumers, their responses may be full probability distributions $p_{c'}(r)$. Throughout, we denote human data by superscript $x$ and synthetic data by $y$.

\subsection{Success Metrics}
We evaluate synthetic panels using two main criteria (see App.~\ref{sec:success-metrics} for more detail):

\textbf{Distributional Similarity.}\  \  
We measure per-survey similarity between synthetic and real purchase intent distributions via Kolmogorov--Smirnov (KS) similarity, i.e.\ $\mathrm{KS\ sim}_s^{xy}=1 - \mathrm{KS\ dist}_s^{xy}$, because it respects the ordinality of the scale. For each experiment, we then report the mean KS similarity $K^{xy}=\mathrm E[\mathrm{KS\ sim}_s^{xy}]$ over all 57 surveys.

\textbf{Relative Concept Appeal and Correlation Attainment.}\  \   
We compute Pearson correlations between mean purchase intents of real and synthetic surveys $R^{xy}=\mathrm{corr}[\mathrm{PI}^x, \mathrm{PI}^y]$ to quantify how well synthetic consumers recover relative concept appeal. Because correlation is upper bounded by noisy human data with a narrow $\mathrm{PI}_s$ distribution, we measure success across all 57 surveys in terms of the maximum attainable correlation, akin to test--retest reliability. For every experiment, we estimate this ceiling by simulating 2,000 test--retest scenarios, splitting each survey into two equally-sized cohorts for each scenario: test and control. Then, the maximum attainable correlation is given by $R^{xx}$ between test and control cohorts. Correlation attainment is then quantified as $\rho = \mathrm{E}[R^{xy}]/\mathrm{E}[R^{xx}]$ where $R^{xy}$ is the correlation between mean purchase intents of the test cohorts and equally-sized random samples of the corresponding synthetic surveys, respectively.

\subsection{Synthetic Response Generation}
Synthetic consumers were instantiated by prompting LLMs with demographic attributes and a product concept (App.~\ref{sec:synthetic-response-generation}). If not stated otherwise we used the full concept image as a stimulus. We evaluated three response strategies (see Fig.~\ref{fig:response-generation-and-esr}A):

\textbf{Direct Likert rating (DLR).}\  \ LLMs reply with a single integer 1, 2, 3, 4, or 5.  

\textbf{Follow-up Likert rating (FLR).}\ \ LLMs first generate a short free-text purchase intent statement, which is then mapped to a Likert score by a new instance of the same model which, this time, received a system prompt to act as a ``Likert rating expert.'' In this system prompt, we included examples of what kind of statements can lead to which rating.

\textbf{Semantic similarity rating (SSR).}\ \  The same free-text responses are embedded and compared to reference statements for each point on the Likert scale, yielding a response probability mass function (pmf) with single probability values being proportional to the cosine similarity between the response and the corresponding reference statement (see Fig.~\ref{fig:response-generation-and-esr}B). Experiments reported in the main text used pmfs that were averaged over six different statement sets for every response (see App.~\ref{sec:reference-statement-sets}). Embedding vectors were retrieved using OpenAI's model ``text-embedding-3-small.''

We used two models (\gpt{} and Gemini-2.0-flash, ``\gem{}'' in the following) and ran experiments with $T_\mathrm{LLM}=0.5$ and $T_\mathrm{LLM}=1.5$. As there was little variation between experiments at different temperatures, we only report results for $T_\mathrm{LLM}=0.5$ in the main text. 

\section{Results}

\subsection{Direct Likert Ratings}

\begin{figure*}
    \centering
    \includegraphics[width=0.92\textwidth]{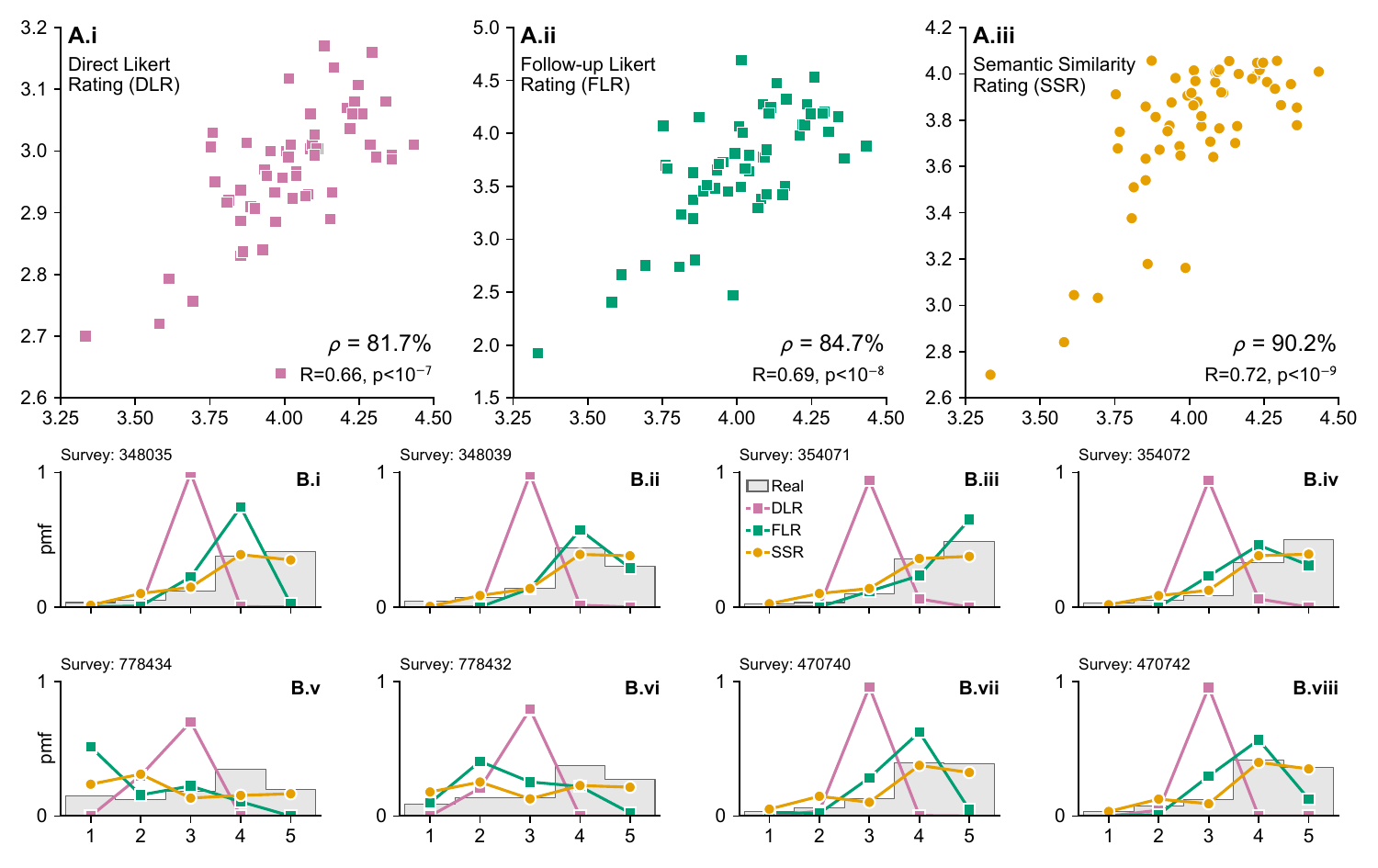} 

    \caption{Comparison of real and synthetic surveys based on \gpt{} with $T_\mathrm{LLM}=0.5$. \textbf{(A)} Mean purchase intent comparison for \textbf{(A.i)} Direct likert ratings (DLRs), \textbf{(A.ii)} textual elicitation with follow-up Likert ratings (FLRs) and \textbf{(A.iii)} semantic similarity ratings (SSRs). \textbf{(B)} Eight example survey response distributions for real surveys and the corresponding synthetic surveys based on DLR, FLR, and SSR, respectively.
    }
    \label{fig:gpt-results}
\end{figure*}

\begin{figure}
    \centering
    \includegraphics[width=0.9\linewidth]{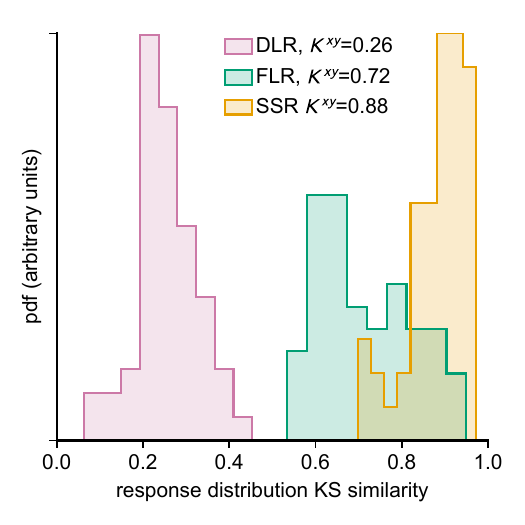} 

    \caption{Comparison of purchase intent distribution similarity between real and synthetic surveys based on \gpt{} with $T_\mathrm{LLM}=0.5$ for direct Likert ratings (DLRs), textual elicitation with follow-up Likert ratings (FLRs) and semantic similarity ratings (SSRs).}
    \label{fig:gpt-results-KS-sim}
\end{figure}

To establish a baseline for comparison, we first tested the performance of asking synthetic consumers for a Likert rating directly, using full information on demographic attributes.
Both LLMs yielded a correlation attainment of about $\rho=80\%$ (cf.\ Fig.~\ref{fig:gpt-results}A.i and Fig.~\ref{fig:gem-results}A.i). At the same time, distributional similarity was poor with a mean similarity of $\Kxy = 0.26$ for \gpt{} and $\Kxy = 0.39$ for \gem{} (cf.\ Figs.~\ref{fig:gpt-results}B, \ref{fig:gpt-results-KS-sim}, ~\ref{fig:gem-results}B, \ref{fig:gpt-results-KS-sim},  and ~\ref{fig:DR-metrics-gpt-T0.5}--\ref{fig:DR-metrics-gem-T1.5}).
Upon detailed inspection of the Likert response distributions, models typically replied with response `3', i.e.~a ``safe'' regression to the center of the scale (cf.\ Figs\ \ref{fig:DR-hists-gpt-T0.5}--\ref{fig:DR-hists-gem-T1.5}). The comparably high correlation with real data was therefore purely a result of occasional responses `2' and `4'. Almost never did the models reply with `1' or `5'. In contrast, the most responses in the real data were values `4' and `5'. Subsequent attempts to nudge models to explore the upper extremes of the distribution via system prompt modification lead to slightly higher distributional similarities while decreasing correlation in mean purchase intent, however. I.e.\ models then ``over-corrected'' in the direction of distributional similarity to real data, resulting in a loss of signal in product ranking, which subverts the overall goal of obtaining useful information about consumer purchase intent.

\subsection{Textual Elicitation with FLR and SSR}

Letting LLMs respond freely and using the generated responses to obtain Likert ratings yields correlation attainment values of $\rho=85\%$ for \gpt{} (see Fig.~\ref{fig:gpt-results}A.ii) and $\rho=90\%$ for \gem{} (see Fig.~\ref{fig:gem-results}A.ii). With \gpt{} consumers, FLRs achieve slightly lower correlation attainment than SSRs with $\rho=90\%$ (see Fig.~\ref{fig:gpt-results}A.iii). With \gem{}, both methods reach similar values (see Fig.~\ref{fig:gem-results}A.iii). 
For the SSR method, distributional similarity markedly increased compared to the naive DLR approach, with $\Kxy=0.88$ for \gpt{} (see Fig.~\ref{fig:gpt-results-KS-sim}) and $\Kxy=0.8$ for \gem{} (see Fig.~\ref{fig:gem-results-KS-sim}). FLRs yield improved distributions compared to DLRs, but fall behind distributional similarity values reached by SSRs, ($\Kxy=0.72$ for \gpt{} and $\Kxy=0.59$ for \gem{}, respectively,  see Figs.~\ref{fig:gpt-results-KS-sim} and \ref{fig:gem-results-KS-sim}). For this rating method, we found that equipping the system prompt with explicit examples of what kind of sentiments may lead to which rating on the Likert scale was necessary to avoid the narrow distributions observed with the DLR approach.

Generally, the synthetic mean purchase intents are far more spread out than the real mean purchase intents: When a product is less attractive, LLMs tend to rate them lower than their human counter parts, on average. For detailed results, see Figs.\ \ref{fig:SSR-hists-gpt-T0.5-00}--\ref{fig:SSR-metrics-gem-T0.5} and Tab.~\ref{tab:model_metrics_all_experiments}.

\subsection{Influence of Demographics and Concept Properties}

\begin{figure}
    \centering
    \includegraphics[width=0.92\linewidth]{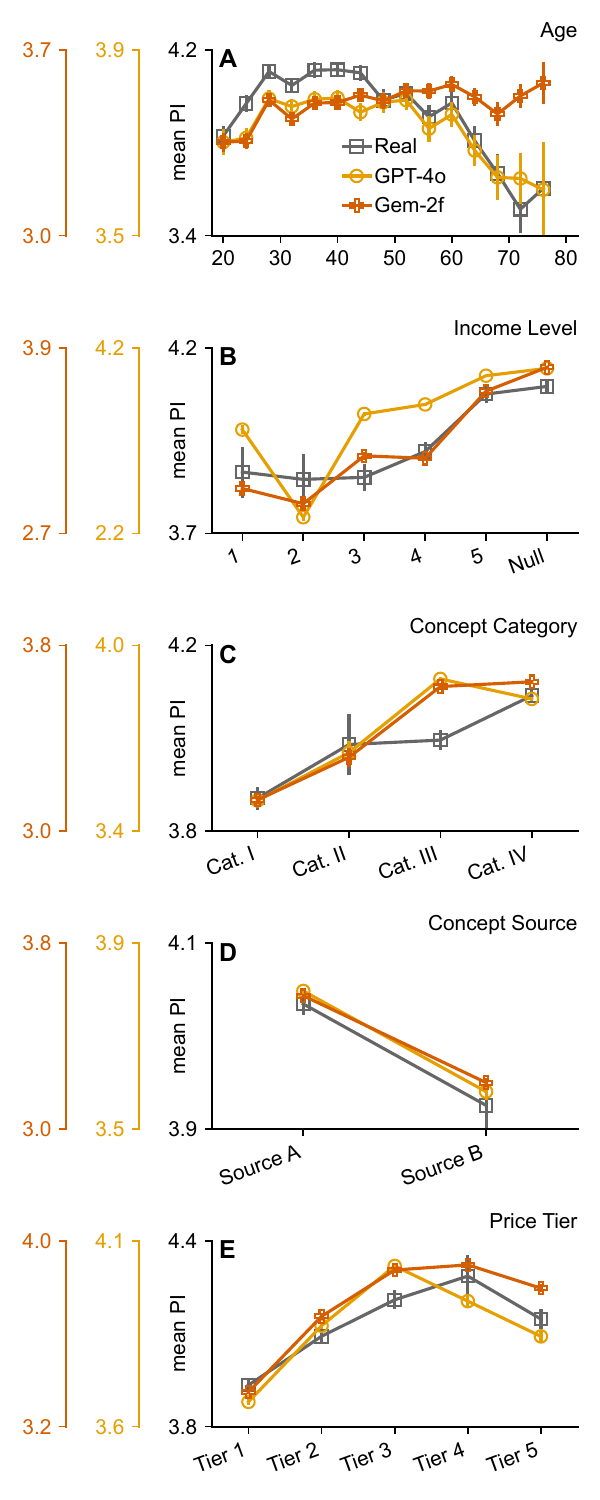} 

    \caption{Mean purchase intent stratified by five demographic and product features (shown are results from the SSR method for both \gpt{} and \gem{}). Error bars represent standard errors.}
    \label{fig:features-infl}
\end{figure}

Furthermore, we are interested to find out which other aspects of the real survey data are mirrored by synthetic consumers (SCs). To this end, we measure mean purchase intent across all products, stratified by demographics and product features and present the results in Fig.~\ref{fig:features-infl}.

Mean purchase intent follows a concave behavior with regards to age: both younger and older participants tended to rate their purchase intent lower than middle-aged age cohorts. This behavior is mirrored by synthetic consumers based on \gpt{}. For \gem{}, younger synthetic consumers reported lower purchase intent. Older consumers, however, reported similar purchase intent as their middle-aged counter parts (see Fig.~\ref{fig:features-infl}A).

In the real surveys, consumers had to rate their income level according to one of six reference statements. Here, statements 1 through 4 all suggested budgetary problems. Hence, it is unsurprising that only for statements 5 and ``Null'' (i.e.~``None of these'') there is a marked increase in purchase intent.
This behavior is replicated both by \gpt{} and \gem{} (see Fig.~\ref{fig:features-infl}B): Personas prompted to have budgetary problems responded with lower purchase intent. \gpt{} reacted very sensitively to being prompted with income level 2, potentially due to the statement's drastic wording of being ``in danger.''

 Both humans and SCs rated ``Cat. IV'' products consistently high and those from ``Cat. I'' consistently the lowest (see Fig.~\ref{fig:features-infl}C).  Moreover, humans and SCs alike reacted negatively to concepts developed by ``Source B'' (see Fig.~\ref{fig:features-infl}D).
Regarding a product's price segment, SCs replicated human behavior rather well once again, rating products from `Tier 3' and `Tier 4' more positively and `Tier 1' lowest (see Fig.~\ref{fig:features-infl}E).

SCs replicated the response behavior less well for gender and dwelling region (see Fig.~\ref{fig:gender-region}). However, mean purchase intent is not being influenced strongly by those features for neither real nor any of the synthetic surveys.

To explore how well models leveraged the information contained in demographic attributes to arrive at the results above, we ran an additional experiment using \gem{} and an SC system prompt that left out all demographic features. 
Surprisingly, this resulted in survey distributions that were very close to the real distributions, with consistently high purchase intent of `4' and `5' and a mean distribution similarity of $\Kxy = 0.91$ (see Figs.~\ref{fig:wodemog-image-metrics-gem-T0.5-00}--\ref{fig:wodemog-image-hists-gem-T0.5-01}). Moreover, we even obtained the same mean and standard deviation as the real data for purchase intent across all surveys $\mathrm {E[PI]}=4.0\pm0.1$. At the same time, for the best reference set, correlation attainment only reached $\rho=50\%$ compared to $\rho=92\%$ for \gem{} SCs prompted with demographic markers, which suggests that if LLMs are not prompted with a detailed enough persona they will rate products more positively in general and will not leverage the actual information in the product concepts enough to produce a meaningful signal. We obtained similar results in further experiments for both models (see Figs.~\ref{fig:demog-hists-gpt-T0.5-00}--\ref{fig:demog-metrics-gem-T0.5}).

\subsection{Additional Results}

While the SSR method is of quantitative nature, the underlying free-text responses make it possible to obtain qualitative feedback on product concepts. Comparing textual responses by humans to those generated by LLMs, we find that the latter are far richer in information, highlight positive features, and raise explicit concerns about less likable product properties. SCs may thereby enrich product research beyond quantitative analyses (see App.~\ref{app:qualitative-evals}).

To test how the SSR method would perform for indicators other than purchase intent, we ran a single experiment for the question ``How relevant was the concept?'' which was posed to the same human participants within each survey.
Taking the average over three new reference sets that were constructed as Likert-scale anchors for this question, we found that the synthetic responses by \gem{} achieved a correlation attainment $\rho=82\%$ for SSR and $\rho=91\%$ for FLRs (cf.\ Fig.~\ref{fig:relevance-metrics-gem-T0.5}). At the same time, the synthetic survey distributions achieved similarity values of $\Kxy = 0.81$ for SSR $\Kxy = 0.62$ for FLRs (cf.\ Figs.\ \ref{fig:relevance-hists-gem-T0.5-00}-\ref{fig:relevance-hists-gem-T0.5-01}), demonstrating the method's potential for generalization.

We further wanted to test how much information the LLMs actually leverage from the product concept beyond coarse-grained features such as demographics and product properties. To this end, we trained 300 LightGBM classifiers on one random half of the studies each and analyzed predicted responses for the other half (see App.\ \ref{app:lightgbm}). 
LightGBM, despite being trained on in-sample data, achieved a correlation attainment of only $\rho=65\%$, compared to $\rho=83\%$ for FLRs and $\rho=88\%$ for SSRs.
Regarding distributional similarity, LightGBM outperformed FLRs with $\Kxy=0.80$ versus $\Kxy=0.72$. However, SSR distributions were still closer to the real survey results with $\Kxy=0.88$, highlighting that zero-shot LLM elicitation---which requires no access to training data from the surveys---synthesizes human responses more effectively than a supervised ML model that does.

To see how a more parsimonious setup would perform, we repeated most of the experiments and replaced the image stimulus with a text stimulus that contained only the product description. We find that doing so mildly reduces the performance for most experiments. Success metrics for all experiments can be found in Tab.~\ref{tab:model_metrics_all_experiments}. 

\section{Discussion and Conclusion}

Our results show that LLM-based synthetic consumers can reproduce core outcomes of traditional consumer concept testing with surprising fidelity. In particular, the semantic similarity rating (SSR) approach yields both realistic distributions of Likert responses and robust product rankings that attain over 90\% of the maximum correlation with human data, based on test--retest reliability. These findings suggest that many of the shortcomings of prior attempts at using LLMs as survey respondents---such as skewed distributions, over-positivity, or regression-to-the-mean---are not intrinsic limitations of LLMs, but rather artifacts of how responses were elicited. By shifting from direct elicitation of Likert responses to textual elicitation and SSR, we resolve many of these artifacts and unlock richer, more interpretable data.

Importantly, no training data or fine-tuning on consumer responses was required. This makes the method widely applicable and inexpensive compared to training or calibration-heavy alternatives. The SSR approach functions as a plug-and-play tool: it translates free-text responses into Likert distributions in a transparent, interpretable way, preserving comparability with canonic survey-based consumer research while also capturing the nuance of unconstrained responses.

A key advantage of this approach is the retention of qualitative information. Whereas human Likert ratings are often accompanied by minimal free-text justifications, LLM-based synthetic consumers readily provide detailed rationales that explain why a product might be attractive or unattractive. These rationales can be mined for themes, objections, or value propositions in ways that raw Likert scores cannot. Moreover, we observe that synthetic responses appear less prone to the positivity bias common in human surveys, producing a wider spread of purchase intent. This broader dynamic range may provide companies with more discriminative signals when evaluating early-stage concepts.

While promising, the method is not without limitations. SSR relies on carefully designed reference statements, and our results show that different anchor sets can lead to slightly different mappings. Averaging across sets mitigates this issue, but future work could optimize reference statements iteratively, or even generate them dynamically with LLMs to maximize alignment with human data. Note that the reference sets created herein were manually optimized for the 57 surveys subject to this study, which means it remains elusive how well they would perform for other surveys. Another limitation concerns demographics: although LLMs captured some demographic patterns (e.g., age and income) quite well, others (e.g., gender, region, ethnicity) were not consistently replicated. This suggests that while persona conditioning does influence synthetic responses, it cannot yet be treated as a reliable proxy for all subpopulations. Researchers must therefore use caution when interpreting subgroup analyses from synthetic panels.

More fundamentally, the usefulness of SSR is bounded by the knowledge domains represented in the LLM’s training data. LLMs are known to hallucinate when asked about unfamiliar topics, and SSR does not eliminate this risk. The reason our approach succeeds in oral care products, and even reflects demographic conditioning, is likely that the model has been exposed to abundant human discussions of these categories in its training corpus (e.g., online forums and consumer reviews). For domains where such background knowledge is sparse or absent, SSR will not conjure valid consumer preferences. Thus, it is important not to view synthetic surveys as universally reliable, but rather as tools whose validity depends on the alignment between training data and the survey domain.

Additionally, SSR's performance depends on the choice of embedding model and similarity measure. Although cosine similarity proved effective, further benchmarking could reveal alternative embedding spaces (e.g., domain-specific encoders) that yield stronger alignment. Finally, while synthetic consumers reproduce human-like distributions and rankings, they cannot fully capture the real-world contingencies of purchasing behavior, such as budget constraints, cultural context, or marketing exposure.

There are several avenues for extending this work. First, the method can be generalized to survey questions beyond purchase intent. By designing reference sets for other Likert constructs such as relevance, satisfaction, or trust we may extend the approach to larger surveys or other applications. Second, optimization strategies could improve SSR: parameters quantifying how a single response is mapped to a distribution could be tuned automatically to maximize correlation with held-out human data. Third, more sophisticated prompting strategies could be explored, such as asking LLMs to generate long free-text responses that are then summarized before mapping to Likert anchors of different questions at once, or multi-stage pipelines in which one LLM generates responses and another critiques or calibrates them. While such methods may be more computationally expensive, they could improve both interpretability and alignment.

Finally, there is an open question about combining SSR with light fine-tuning approaches. Although we deliberately avoided training data here to demonstrate generality, hybrid methods where SSR is used in tandem with calibration or prompt optimization may achieve even higher fidelity. Crucially, however, SSR provides a low-cost baseline that narrows the gap between synthetic and human survey data without requiring retraining.

If further validated, SSR-enabled synthetic consumers could substantially change how early-stage product research is conducted. Instead of commissioning large human surveys for every product idea, companies could first screen concepts synthetically, reserving human panels for the most promising candidates. This would reduce costs, accelerate iteration, and enable smaller firms to access consumer insights that were previously out of reach. At the same time, the availability of detailed synthetic rationales could complement human panels, offering a richer understanding of consumer perceptions.

In summary, by combining interpretability, statistical reliability, and qualitative richness, SSR addresses many of the challenges that have constrained the use of LLMs as synthetic survey respondents. While not a wholesale replacement for human research, SSR establishes a credible framework for augmenting and accelerating consumer insight generation.

\begin{acks}
The original manuscript was written by all authors, ChatGPT-5 was subsequently used in all sections to reformulate and condense the text.
\end{acks}

\newpage

\appendix

\section*{Appendix}

\section{Detailed Methods}
\label{app:methods}

\subsection{Data}
\label{app:data}

We base our study on a corpus of 57 consumer research surveys, conducted using a digital consumer research platform, supplied by a leading personal care corporation.
Each survey relates to a unique hypothetical personal care product concept designed for the US market.
In the dataset, a product concept is represented on a presentation slide that contains at minimum a text description and in many instances a concept image, as well.
Every survey had a unique set of participants, with sizes ranging from $N_s=150$ to $N_s=400$.
In total, the corpus has 9,300 unique participants.
For most of the surveys, age, gender, and location of the participants is known.
To a lesser extent, income level is reported, and only nine surveys contain information on consumer ethnicity.

While surveys prompt participants to score product concepts along various dimensions, we focus on the central question of \textit{purchase intent}, often phrased as ``Based on everything you've seen and heard, how likely are you to purchase the product?''.
This response was scored on a 5pt Likert scale, requiring participants to select one of the integer numbers $i=1, 2, 3, 4, 5$.

\subsection{Definitions}
\label{app:definitions}

Let the corpus of all surveys be called $\mathcal{S}$.
A single survey $s\in \mathcal{S}$ consists of:
\begin{enumerate}
    \item $N_s=|\mathcal{C}_s|$ consumers $c\in \mathcal{C}_s$ with demographic attributes $\bm d_c = \{d_{c,1}, d_{c,2}, \ldots, d_{c,D}\}$ containing features such as age, gender, income, location, and ethnicity, as well as
    \item a single product concept.
\end{enumerate}
Note that we need not formally distinguish between product concepts and surveys as each survey only relates to a single unique concept, hence $s$ may either denote a survey or a concept depending on the context.

Let $r_{c}$ be the Likert scale rating response provided by a human consumer $c$ after reviewing a product concept $s$, and asked about their purchase intent (since there are no consumers that repeat across surveys, we need not index the response by $s$). Response $r_c$ may be equal to any of the integer numbers $i=1, 2, 3, 4, 5$.
Having produced a Likert scale response, each consumer $c\in\mathcal C_s$ of a given survey $s$ is associated with a response probability mass function (pmf) of $p_{c}(i) = \delta_{ir_c}$, where $\delta_{ir_c}$ is the Kronecker delta function.
The whole survey distribution, which aggregates responses from all consumers, is then given by:
\begin{align}
    p_s(i) = \frac{1}{N_s} \sum_{c\in\mathcal C_s} \delta_{ir_c}
\end{align}
The mean purchase intent of the concept $s$ is then calculated as:
\begin{align}
    \mathrm{PI}_s = \sum_{i=1}^5ip_s(i).
\end{align}

We describe as a \emph{synthetic consumer} $\tilde c$ an LLM that was prompted to impersonate a human consumer with demographic attributes $\bm d_c$ or a subset thereof, see Section \ref{sec:synthetic-response-generation}.

All of the definitions outlined above apply for synthetic consumers $\tilde{c}$, as well, however with the important distinction that we do not restrict those to reply with single-response distributions, i.e. we do not require that the response pmf is a Kronecker delta function.
Instead, a synthetic consumer response may yield an arbitrary pmf $p_{\tilde{c}}(i)$ on the Likert scale.
As we shall see in Section \ref{sec:synthetic-response-generation}, this involves mapping a textual response $t_{\tilde{c}}$ to a Likert scale rating $r_{\tilde{c}}$ integer, or a pmf $p_{\tilde{c}}(i)$.

Henceforth, we will denote real data with the superscript $x$ and synthetic data with the superscript $y$. 

\subsection{Success Metrics \label{sec:success-metrics}}
We define two success metrics, one to measure the distributional similarity between outcomes of synthetic and real surveys, and another to measure the degree to which synthetic consumers replicate the concept ranking obtained from real surveys.

\paragraph{Distributional Similarity}

Synthetic consumer panels should replicate real consumer purchase intent distributions as accurately as possible.
To this end, we define the distributional similarity between a synthetic and real survey $s$ as the complement of the Kolmogorov-Smirnov (KS) distance:
\begin{align}
    \mathrm{KS\ sim}_s^{xy} = 1- \sup_{r=1,\dots,5}|F_s^x(r) -F_s^y(r)|.
\end{align}
Likert data responses are ordinal and because there is no measurable concept of distance between the integer responses, technically, the (KS) distance is an inappropriate measure.
However, we find that in practice KS distance has various advantages, for instance it is simple to interpret, as the maximum distance between two CDFs which is always a number between 0 (where distributions are equal) and 1 (where distributions are entirely dissimilar).
Second, the ordinality of the data is respected, i.e.\ it does make a strong difference whether two distributions have peaks that lie close or far away from each other.

At times, we will compare KS similarity to distributional cosine similarity defined as
\begin{align}
    C_s^{xy} =
    \frac{\bm p_s^x\cdot \bm p_s^y}
        {|\bm p_s^x| |\bm p_s^y|}
\end{align}
which does not respect the scale's ordinality. Here, we treat the pmf as a vector $\bm p = (p(1), \dots, p(5))$.

We denote as $\Kxy = \mathrm E [\mathrm{KS\ sim}_s^{xy}]$ the mean distributional similarity over all surveys (and analogously, $C^{xy} = \mathrm E[C_s^{xy}]$).

\paragraph{Concept Ranking Similarity and Correlation Attainment}

A concept's popularity in terms of mean purchase intent should rank similarly for both synthetic surveys as well as real surveys.
To measure how similarly concepts are perceived, we compute the Pearson correlation between the mean purchase intents from synthetic and real surveys $\mathrm{PI}^y$ and $\mathrm{PI}^x$, respectively:
\begin{align}
    R^{xy} = \mathrm{corr}[\mathrm{PI}^x, \mathrm{PI}^y].
\end{align}

Naively, we should thus expect perfect synthetic consumers to yield $R^{xy}$ close to 1.
However, since we observe that the mean purchase intents of real surveys lie relatively close to each other with $\mathrm E[\mathrm{PI}^x]=4.0$ and $\mathrm{Std}[\mathrm{PI}^x]=0.2$, we must consider the possibility that these results are influenced by noise and hidden biases in a non-negligible manner.
We therefore ask: were the surveys repeated with a new cohort of similarly drafted real consumers, how well would the mean purchase intent of the repeated surveys correlate with the mean purchase intent of the original surveys?
This value, $R^{xx}$, should be the theoretical maximum concept ranking similarity we judge the performance of the synthetic consumers by.
In other words, a high concept ranking correlation between synthetic and real consumers is one that approaches the retest correlation of real consumer responses. 

Although we cannot obtain a traditional test--retest reliability estimate by repeating each survey with a new cohort of real consumers of size $N_s$, we can simulate retest scenarios a large number of times where we randomly split survey participants into test and control cohorts of size $N_s/2$.

To obtain a reliability measure of the concept ranking, we perform the following comparison:
For every survey $s$, we split the participant set $\mathcal{C}_s$ in half at random.
One half $\mathcal{C}_{s,t}$ will be called the \emph{test cohort}, whose responses constitute the central survey results.
We call the second half $\mathcal{C}_{s,c}$ the \emph{control cohort}, whose responses represent a repeated survey to control and compare the results of the first survey.
Then, for the corresponding synthetic survey with participants $\mathcal{C}_s^y$, we randomly sample a test cohort of the same size as the corresponding real test and control cohorts, respectively.
We follow this cohort construction procedure once for every survey, achieving corresponding average purchase intents $\mathrm{PI}_{s,t}^x$, $\mathrm{PI}_{s,c}^x$, and $\mathrm{PI}_{s,t}^y$, such that correlation coefficients $R^{xx} = \mathrm{corr}\left[\mathrm{PI}_{s,t}^x, \mathrm{PI}_{s,c}^x\right]$ and $R^{xy} = \mathrm{corr}\left[\mathrm{PI}_{s,t}^x, \mathrm{PI}_{s,t}^y\right]$ can be computed.
Repeating this experiment $m=2\,000$ times and taking the average over the respective correlation coefficient, we obtain \emph{correlation attainment}
\begin{align}
    \rho = \frac {\mathrm E[R^{xy}]} {\mathrm E[R^{xx}]},
\end{align}
quantifying how close the correlation coefficient between real and synthetic consumers is to the theoretical maximum.

\subsection{Synthetic Response Generation \label{sec:synthetic-response-generation}}

For every human consumer $c$, a \textit{synthetic consumer} $\tilde{c}$ was constructed by priming an LLM to be a participant in a product research survey and to impersonate a consumer $c$ with the same or a subset of demographic attributes $\bm d_{\tilde{c}} \subseteq \bm d_c$.
Then the LLM was shown the product concept, in the form of an image containing the text and potentially an image of the product (see App. \ref{sec:product-concept-examples}).
Subsequently, the LLM was prompted with the question ``How likely are you to purchase the product?'', and a response was sampled.
Due to the stochastic nature of LLMs, we designed our experimental setup such that any number of repeat samples $n$ could be drawn upon submitting each prompt, enabling us to average results over multiple samples.
In the following analysis we use $n=2$ samples, which we found was sufficient to obtain stable results.

We focused on models by Google and OpenAI; after initial experiments with different models such as ```gemini-1.5-flash,'' ``gemini-2.5'', and ``o3'' we settled on ``gemini-2.0-flash'' (referred to as ``\gem'' throughout the text) and ``gpt-4o'' (``\gpt{}'') for production runs as these models gave the most consistent responses across all experiment types.
Experiments were run with parameters $p_\mathrm{top}=0.9$ and temperature $T_\mathrm{LLM}\in\{0.5,1.5\}$ if not noted otherwise.

Below we describe different approaches to generate synthetic responses.

\subsubsection{Direct Likert Rating.}
\label{sec:direct-likert}

The simplest approach to generate a Likert scale rating from an LLM which has been presented with a product concept, is to treat the LLM like a human participant and let it respond with a single token, i.e.\ one of the integer responses 1, 2, 3, 4, or 5.
This approach is straightforward and produces results with minimal effort.

\subsubsection{Follow-up Likert Rating (Textual Elicitation Before Rating)}
\label{sec:follow-up-likert}

A slightly more sophisticated approach is to let the LLM first express its liking of the product concept in a brief but otherwise unconstrained text response $t_{\tilde{c}}$, and only afterward let it summarize this in a single integer response $r_{\tilde{c}}$.
Specifically, after priming the LLM with its demography and showing a product concept, we prompt it with the question ``How likely are you to purchase the product?'', and stating ``Reply briefly to any questions posed to you.'' in the system prompt. After sampling the response, we prompt the LLM to be a ``Likert rating expert'' and request it to map the text response it just gave, to the corresponding integer response on a 5pt Likert scale. The same LLM that generates a response is also tasked with rating it, using $p_\mathrm{top}=1$ and $T_\mathrm{LLM}=0.3$.

\subsubsection{Semantic Similarity Rating (SSR)}
\label{sec:esr}

Mapping textual responses to Likert scale ratings is, however, non-trivial, as a response rarely maps exactly to one and only one rating.
For instance, a response may read ``I'd probably buy it. I like that it's easy to use and I can take it with me. Plus, the price isn't too bad.''
Depending on how the scale is defined and who is on the receiving end of this statement, the response most likely would map to a ``4'' or a ``5''.
The statement could be easily interpreted as purchase being very likely (5) when imagining a casual conversation about purchase intent.
Others might interpret this response as just ``likely'' (4).
Hence, there's an inherent ambiguity in the textual responses that gets discounted through the mapping onto a single number.

We therefore propose an alternative procedure that maps a textual response to a distribution of Likert scale ratings.
To do this, we construct reference statements $\sigma_{r}$ that each map to a Likert scale rating, then estimate the similarity of the response text $t_{\tilde{c}}$ to each of these reference statements, and use the similarities to construct a pmf of Likert scale ratings.
In practice, we construct $m$ such reference statement sets $\Sigma_i$ where $i=0,\dots,m-1$, across which we eventually average the respective single-response pmfs to obtain a single-response result pmf.
In this analysis, we use $m=6$ sets (see App.~\ref{sec:reference-statement-sets}).
They are all similar but not identical, and designed to capture the different ways a consumer may express their purchase intent.
However, as this use of multiple sets of reference statements is more an implementation detail than a theoretical innovation, and since it will complicate the mathematical notation unnecessarily, we do not explicitly mention it in the following.

To effectively compute the similarity between a response text $t_{\tilde{c}}$ and a reference statement $\sigma_r$, we retrieve embedding vectors $\bm v_{\sigma_{r}}$ and $\bm v_{t_{\tilde{c}}}$ from a text-embedding model for each of the reference statements as well as the response text from synthetic consumer $\tilde{c}$, respectively. In this study, we exclusively use OpenAI's ``text-embedding-3-small'' model. Tests with ``text-embedding-3-large'' left the results virtually unchanged.
With these vectors, every response can be mapped to a similarity in the embedding space by means of cosine similarity as
\begin{align}
    \gamma(\sigma_r, t_{\tilde{c}}) = 
    \frac{
    \bm v_{\sigma_{r}}\cdot\bm v_{t_{\tilde{c}}}
    }{
                               |\bm v_{\sigma_r}|     |\bm v_{t_{\tilde{c}}}|
    }.
\end{align}

We then interpret this similarity as proportional to a response probability $p_r$ for integer response $r$, such that
\begin{align}
    p_{\tilde{c},i}(r) \propto 
    \gamma(\sigma_{r,i}, t_{\tilde{c}})-\gamma(\sigma_{\ell,i}, t_{\tilde{c}}) + \epsilon \delta_{\ell,r}
\end{align}
where $\ell$ is the reference statement with minimum similarity over the corresponding set $\Sigma_i$ and  normalization $\sum_{r=1}^5 p_{\tilde c, i}(r)=1$.
Note that subtracting the minimum similarity over the reference statement set $\gamma(\sigma_{\ell,i}, t_{\tilde{c}})$ is a way to adjust for potential low variance in the similarity scores.
In practice, we observe that within the space of all embeddable language, the numerical difference between extremes like ``It's very unlikely that I'd buy it.'' and ``It's very likely that I'd buy it.'' is numerically small, and so if we do not subtract the minimum similarity, the resulting pmf will be almost entirely flat.
For $\epsilon=0$, this equation can be read as follows: For every similarity, subtract the minimum similarity over the reference statement set.
Normalize the remaining similarities by the total sum to obtain a probability mass function (pmf).
Of course, following this procedure means that for every textual response we obtain a rating distribution where one of the ratings has zero likelihood to occur.
The parameter $\epsilon$ offsets this bias and makes it more controllable.

To make this mapping from a similarity to a probability mass function (pmf) even more controllable, we can introduce a tempe\-rature-like parameter that controls how ``smeared out'' the resulting pmf should be
\begin{align}
\label{eq:esr-temperature}
    p_{\tilde{c},i}(r,T) \propto 
    p_{\tilde{c},i}(r)^{1/T}
\end{align}
with $\sum_{r} p_{\tilde c, i}(r,T)=1$.

While we restrict our study to $\epsilon=0$ and $T=1$, it is worth introducing them as levers to make the SSR mapping more controllable.
One can, for instance, design an optimization procedure to find values for $\epsilon$ and $T$ that yield the best SSR mapping in terms of the success metrics defined in Section \ref{sec:success-metrics}. A first test suggests that $T=1$ is a reasonable choice as a rule of thumb, but that there is optimization potential: clearly there are optima to be found around $T=1$ both for maximizing correlation and distribution similarity (see  Fig.~\ref{fig:temperature-scan-SSR-gpt-T0.5}).
A full Python code implementation of SSR is given in Appendix \ref{sec:esr-implementation}.

\section{Product Concept Example \label{sec:product-concept-examples}}

\begin{figure*}[thp]
    \includegraphics[width=\textwidth]{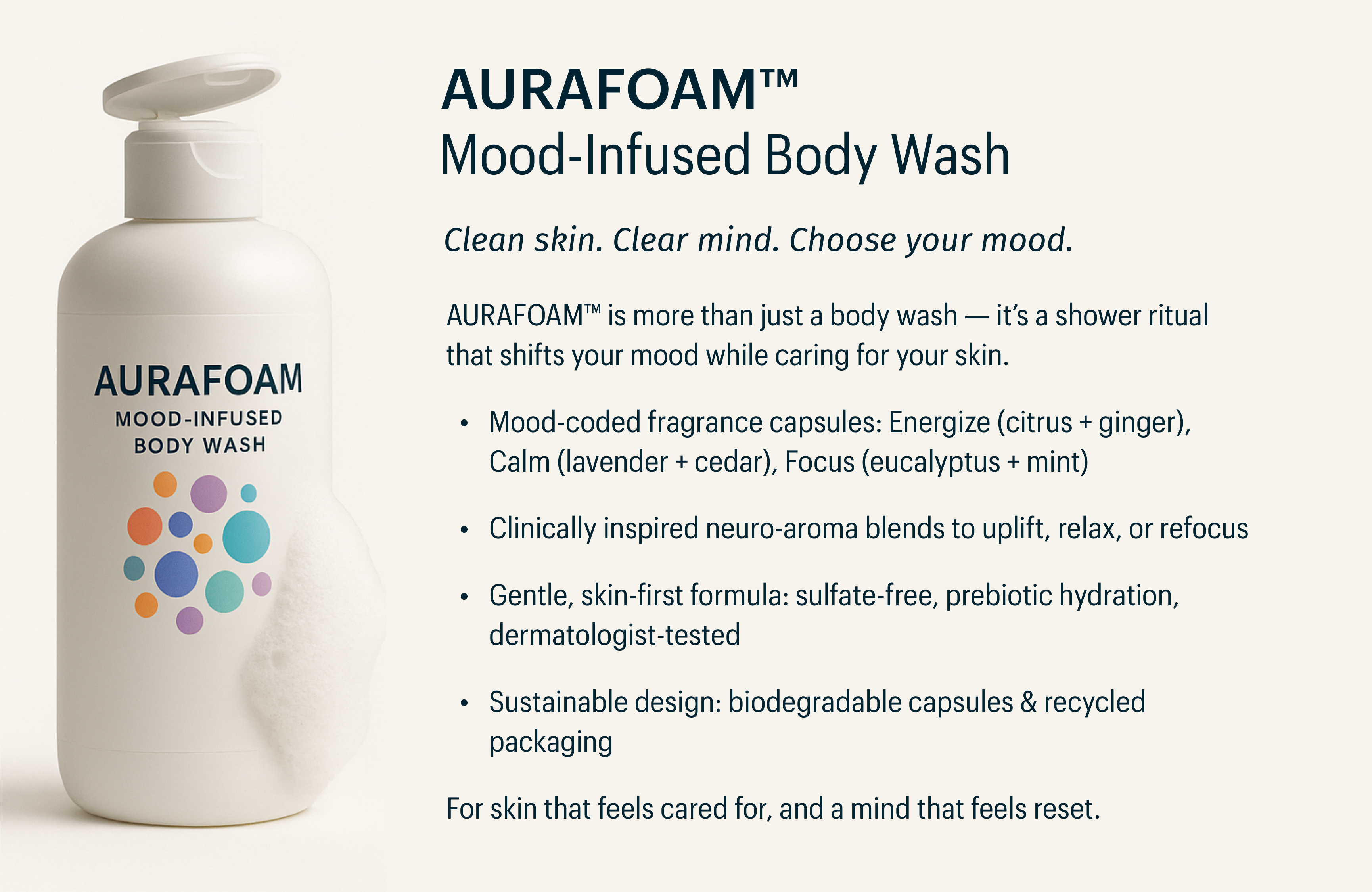} 

    \caption{A surrogate product concept similar to those used in the 57 concept surveys. }
    \label{fig:productexample}
\end{figure*}

Fig.\ \ref{fig:productexample} shows an examples of a product concept image similar to those used in the study. When we refer to ``image stimulus'' in the main text, an image like this, including either both an illustration and the concept description or only a concept description was supplied to an LLM synthetic consumer. When we refer to ``text stimulus,'' only the text was given to the synthetic consumer. For the latter, the text was previously transcribed from the product concept image using \gpt{}. 

\section{Additional Material for SSR}

\subsection{Reference Statement Sets \label{sec:reference-statement-sets}}

To map free-text responses onto a 5pt Likert scale, we constructed six sets of anchor statements, five statements in each set (one for each Likert category 1 through 5). These anchors serve as semantic prototypes against which model-generated responses are compared in embedding space. Each anchor statement was written to reflect the intensity of purchase intent associated with its corresponding Likert rating:
\begin{itemize}
    \item The lowest anchor expresses a purchase to be unlikely.
    \item The middle anchor reflects indifference or uncertainty.
    \item The highest anchor conveys intent or possible intent to purchase.
\end{itemize}
The remaining two intermediate anchors were formulated as semantically in between their adjacent statements.
The anchors were designed to be short, generic, and domain-independent, such that they could plausibly apply to any consumer product concept. Their purpose is not to capture the nuances of any specific product, but to provide a reference gradient of intent from ``purchase improbable'' to ``purchase probable.'' This approach allows defining the Likert scale in a way that adapts to the stylistic tendencies of LLM responses, ensuring that the full range of intent is captured.

\subsection{SSR Implementation \label{sec:esr-implementation}}

We give a full Python implementation of the SSR method, available on GitHub at \url{https://github.com/pymc-labs/semantic-similarity-rating}.
The package offers a simple programming interface for generating SSR-based Likert scale response distributions from LLM responses.

\section{Machine-Learning Comparison Based on LightGBM}
\label{app:lightgbm}

To benchmark the performance of zero-shot LLM elicitation against a classical machine learning approach, we trained gradient-boosted decision tree models (LightGBM \cite{lightgbm}) on subsets of the survey data. The goal was to assess how well a model trained on demographic and product features could reproduce individual Likert ratings compared to synthetic responses generated by LLMs.

We considered the complete set of 57 consumer concept surveys used in the main study. For each of 300 iterations, we randomly split the surveys in half. One half (28 surveys) was used for training, the other half (29 surveys) for testing. This setup mirrors the cross-study generalization scenario relevant for real-world applications, where new concepts and respondents are unseen during training.

On each training split, we fitted a LightGBM classifier with the following input features:
\begin{itemize}
    \item \textbf{Demographics (5 features):} age, gender, income tier, region, and ethnicity.
    \item \textbf{Concept attributes (3 features):} category of personal care, price tier, and concept source.
\end{itemize}
The target variable was the purchase-intent rating on a 5-point Likert scale. Models were trained with default LightGBM hyperparameters, using multiclass classification (with the five Likert values treated as separate classes). Missing feature data was assigned a ``Null'' category.

For each held-out study, we predicted Likert responses for all respondents and aggregated them into synthetic response distributions. We then compared these predictions to the original human survey data using two metrics: (i) test--retest reliability ($\rho$), defined analogously to the procedure in the main text, i.e. by Pearson correlation between synthetic and human mean purchase intents of the 29 surveys of the test set, split in half once again, and (ii) distributional similarity ($\Kxy$), defined as the complement of the Kolmogorov--Smirnov distance between the predicted and observed Likert distributions for the 29 surveys in the test set.

Across 300 iterations, LightGBM achieved a correlation attainment of $\rho = (64.6 \pm 1.0)\%$, substantially below both follow-up Likert ratings ($83.2 \pm 0.7\%$) and SSR-based elicitation with \gpt{} at $T_\mathrm{LLM}=0.5$ and image stimulus ($88.3 \pm 0.7\%$). For distributional similarity, LightGBM ($\Kxy = 0.797 \pm 0.002$) outperformed follow-up Likert ratings ($0.716 \pm 0.001$) but was surpassed by SSR ($0.883 \pm 0.001$).

This analysis shows that a simple supervised ML model trained on demographic and product features cannot match the fidelity of zero-shot LLM elicitation in recovering human-like response behavior. While LightGBM achieves moderate distributional alignment, its markedly lower reliability underscores the advantage of LLM-based methods that leverage semantic understanding of product descriptions without requiring additional training data.

\section{Textual Responses Allow for Qualitative Evaluations}
\label{app:qualitative-evals}

As a byproduct of applying the SSR method to obtain quantitative ratings, the textual responses generated by LLMs are rich in information and make it possible to evaluate the product concept in more detail.

While the real surveys focused on ratings on a Likert scale, they also asked people to write free text as responses to the questions ``What did/didn't you like about the concept?'' The replies to these open questions are lacking depth and only seldom provide important feedback. Typically, they are rather short like ``It's good'', or just repeat information contained in the concept like ``Not much, just the steps and how it tells you what it was for.'' Less frequently, participants gave actual feedback about what they liked, for instance one such response reads ``Inexpensive and affordable. New \& light. [Application] from your own home.'' In contrast, the replies about purchase intent by synthetic consumers provide much more in-depth feedback about why or why not they would possibly purchase the product. For instance, one synthetic consumer wrote: ``The ease of use and [...] safety are appealing, but I'd want to know more about its effectiveness and any potential side effects.'' Another responded: ``The ease of use and the promise of no [...] sensitivity are appealing. Plus, it's from a trusted brand.''

Similarly, synthetic consumers rarely held back with their criticisms, which, at times, came written in the style of the persona they were asked to imitate. For a less positively received concept, \gpt{} synthetic consumers responded ``It seems a bit too high-end for my needs and budget'' and ``[These body parts] don't usually bother me, so I don't think I need it'' while two based on \gem{} wrote e.g.,\ 
``Seems kinda bougie for [this kind of product]''
and
``Sounds expensive, and I'm not sure I buy all that `microbiome' talk. I'll stick with what I know'', respectively.

In total, the responses generated by LLMs can be leveraged to obtain detailed feedback on why or why not a product concept was rated with higher or lower purchase intent. Additionally, synthetic consumers seem to suffer less from a positivity bias, as demonstrated by the wider spread of mean purchase intent measured in the previous section as well as confirmed by the qualitative analysis of synthetic responses.

\begin{table*}[t]
\centering
\caption{Metrics for all experiments on purchase intent. ``Direct'' refers to direct Likert rating elicitation (DLR), ``Textual'' refers to free-text responses followed by SSR and FLR. We show correlation attainment $\rho$, distributional similarities $\Kxy$ and $C^{xy}$, mean purchase intent correlation $R^{xy}$ between human and synthetic surveys, including results for best-comparison set $\Sigma$ (see App.~\ref{sec:reference-statement-sets}). We also report mean survey purchase intent and its standard deviation.}
\label{tab:model_metrics_all_experiments}
\begin{tabular}{l l l l l c | c c | c c | c c | c c | c c }
\hline
\hline
\textbf{Elicit.} & \textbf{Dem.} & \textbf{Model} & \textbf{Stim.} &\textbf{$T_{\mathrm{LLM}}$} & \textbf{Best} $\Sigma$   & \multicolumn{2}{c|}{$\rho$ (\%)} & \multicolumn{2}{c|}{$\Kxy$} & \multicolumn{2}{c|}{$R^{xy}$} & \multicolumn{2}{c|}{$C^{xy}$} & \multicolumn{2}{c}{$\mathrm {E[PI_s]}\pm\mathrm{std}$} \\
 & & & & & & \textbf{SSR} & \textbf{Lik.}  & \textbf{SSR} & \textbf{Lik.}  & \textbf{SSR} & \textbf{Lik.} & \textbf{SSR} & \textbf{Lik.} & \textbf{SSR} & \textbf{Lik.} \\
\hline
                Direct  & Full & \gpt{} & Text  & 1.5 &      &      & 88.5 &      & 0.37 &       & 0.717 &      & 0.39 &                 & 2.95 $\pm$ 0.44 \\
                Direct  & Full & \gpt{} & Text  & 0.5 &      &      & 89.6 &      & 0.36 &       & 0.720 &      & 0.38 &                 & 2.96 $\pm$ 0.45 \\
                \hline
                Direct  & Full & \gpt{} & Image & 1.5 &      &      & 78.7 &      & 0.29 &       & 0.648 &      & 0.29 &                 & 2.97 $\pm$ 0.16 \\
                Direct  & Full & \gpt{} & Image & 0.5 &      &      & 81.7 &      & 0.26 &       & 0.661 &      & 0.26 &                 & 2.96 $\pm$ 0.11 \\
                \hline
                Direct  & Full & \gem{} & Text  & 1.5 &      &      & 68.4 &      & 0.46 &       & 0.546 &      & 0.48 &                 & 3.28 $\pm$ 0.50 \\
                Direct  & Full & \gem{} & Text  & 0.5 &      &      & 67.5 &      & 0.45 &       & 0.541 &      & 0.47 &                 & 3.28 $\pm$ 0.50 \\
                \hline
                Direct  & Full & \gem{} & Image & 1.5 &      &      & 82.5 &      & 0.41 &       & 0.660 &      & 0.41 &                 & 3.17 $\pm$ 0.40 \\
                Direct  & Full & \gem{} & Image & 0.5 &      &      & 80.2 &      & 0.39 &       & 0.640 &      & 0.40 &                 & 3.17 $\pm$ 0.40 \\
                \hline
                Textual & Full & \gpt{} & Text  & 1.5 & 4    & 85.0 &      & 0.85 &      & 0.691 &       & 0.94 &      & 3.75 $\pm$ 0.40 &      \\
                Textual & Full & \gpt{} & Text  & 0.5 & 4    & 83.1 &      & 0.84 &      & 0.680 &       & 0.92 &      & 3.71 $\pm$ 0.45 &      \\
                Textual & Full & \gpt{} & Text  & 0.5 & avg. & 82.5 & 69.2 & 0.84 & 0.64 & 0.675 & 0.562 & 0.94 & 0.70 & 3.63 $\pm$ 0.42 & 3.39 $\pm$ 0.79 \\
                \hline
                Textual & Full & \gpt{} & Image & 0.5 & 1    & 91.9 &      & 0.82 &      & 0.740 &       & 0.93 &      & 3.67 $\pm$ 0.36 &      \\
                Textual & Full & \gpt{} & Image & 0.5 & avg. & 90.2 & 84.7 & 0.88 & 0.72 & 0.724 & 0.687 & 0.96 & 0.80 & 3.77 $\pm$ 0.31 & 3.67 $\pm$ 0.55 \\
                \hline
                Textual & Full & \gem{} & Text  & 1.5 & 3    & 76.3 &      & 0.81 &      & 0.611 &       & 0.92 &      & 3.56 $\pm$ 0.40 &      \\
                Textual & Full & \gem{} & Text  & 0.5 & 3    & 72.7 &      & 0.81 &      & 0.581 &       & 0.91 &      & 3.56 $\pm$ 0.42 &      \\
                Textual & Full & \gem{} & Text  & 0.5 & avg. & 73.0 & 73.5 & 0.80 & 0.62 & 0.583 & 0.589 & 0.91 & 0.69 & 3.52 $\pm$ 0.45 & 3.57 $\pm$ 0.86 \\
                \hline
                Textual & Full & \gem{} & Image & 0.5 & 5    & 92.4 &      & 0.82 &      & 0.737 &       & 0.92 &      & 3.64 $\pm$ 0.45 &      \\
                Textual & Full & \gem{} & Image & 0.5 & avg. & 90.6 & 92.1 & 0.80 & 0.59 & 0.720 & 0.736 & 0.91 & 0.64 & 3.51 $\pm$ 0.42 & 3.33 $\pm$ 0.75 \\
                \hline
                Textual & None & \gpt{} & Text  & 1.5 & 4    & 50.1 &      & 0.92 &      & 0.409 &       & 0.98 &      & 4.05 $\pm$ 0.12 &      \\
                Textual & None & \gpt{} & Text  & 0.5 & 4    & 45.3 &      & 0.91 &      & 0.390 &       & 0.97 &      & 4.09 $\pm$ 0.16 &      \\
                Textual & None & \gpt{} & Text  & 0.5 & avg. & 47.4 & 41.2 & 0.91 & 0.57 & 0.408 & 0.324 & 0.98 & 0.77 & 3.92 $\pm$ 0.14 & 4.61 $\pm$ 0.51 \\
                \hline
                Textual & None & \gem{} & Text  & 1.5 & 3    & 60.5 &      & 0.87 &      & 0.481 &       & 0.95 &      & 3.85 $\pm$ 0.24 &      \\
                Textual & None & \gem{} & Text  & 0.5 & 3    & 49.5 &      & 0.87 &      & 0.411 &       & 0.95 &      & 3.90 $\pm$ 0.26 &      \\
                Textual & None & \gem{} & Text  & 0.5 & avg. & 57.4 & 58.0 & 0.91 & 0.62 & 0.481 & 0.480 & 0.97 & 0.75 & 3.90 $\pm$ 0.26 & 4.26 $\pm$ 0.58 \\
                \hline
                Textual & None & \gem{} & Image & 0.5 & 4    & 50.1 &      & 0.91 &      & 0.414 &       & 0.98 &      & 4.09 $\pm$ 0.07 &      \\
                Textual & None & \gem{} & Image & 0.5 & avg. & 15.5 & 64.3 & 0.91 & 0.67 & 0.143 & 0.530 & 0.98 & 0.78 & 3.94 $\pm$ 0.08 & 4.25 $\pm$ 0.34 \\
                \hline
\end{tabular}
\end{table*}

\begin{figure*}
    \centering
    \includegraphics[width=0.92\textwidth]{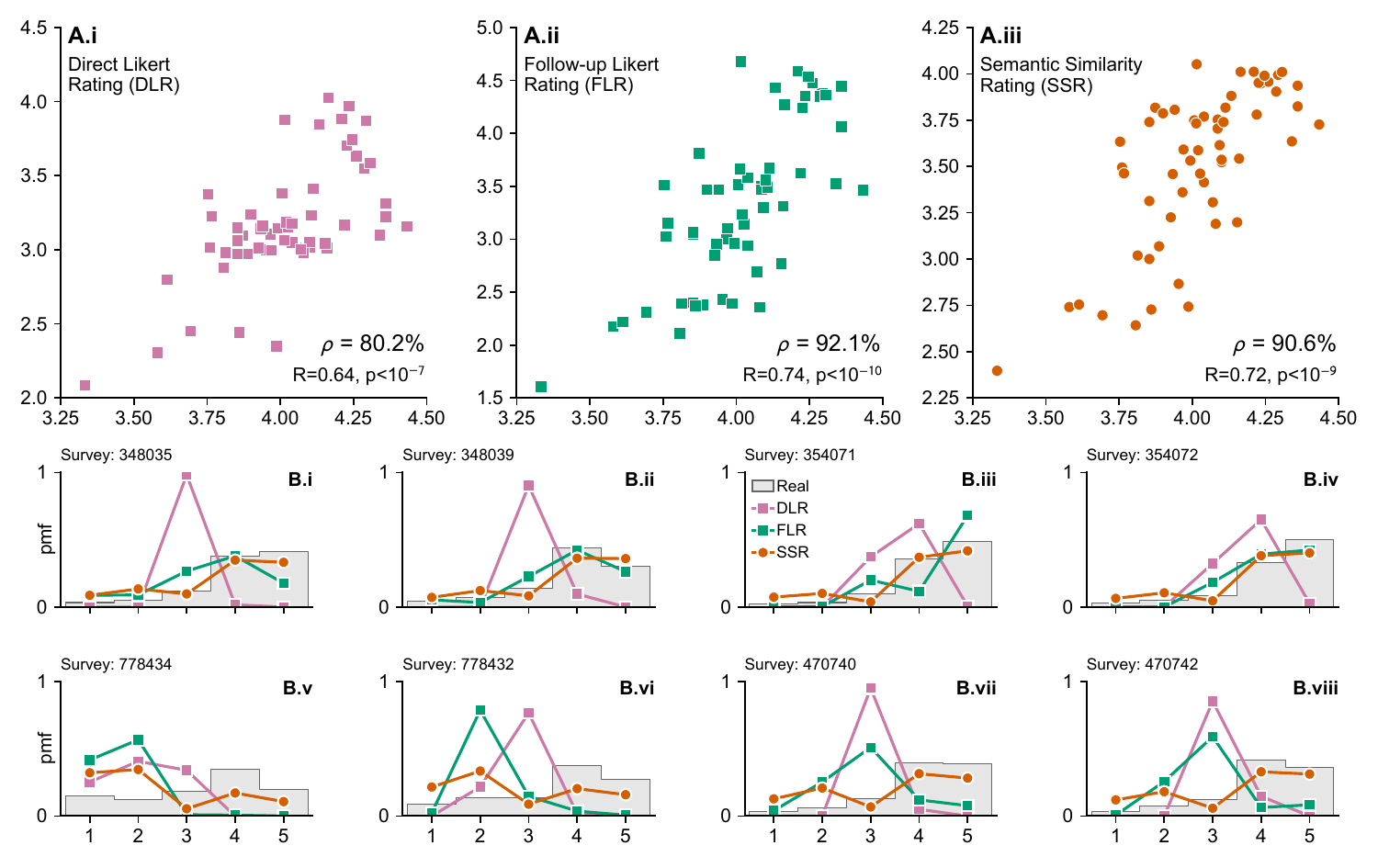} 

    \caption{Comparison of real and synthetic surveys based on \gem{} with $T_\mathrm{LLM}=0.5$. \textbf{(A)} Mean purchase intent comparison for \textbf{(A.i)} Direct likert ratings (DLRs), \textbf{(A.ii)} textual elicitation with follow-up Likert ratings (FLRs) and \textbf{(A.iii)} semantic similarity ratings (SSRs). \textbf{(B)} Eight example survey response distributions for real surveys and the corresponding synthetic surveys based on DLR, FLR, and SSR, respectively.
    }
    \label{fig:gem-results}
\end{figure*}

\begin{figure*}
    \centering
    \includegraphics[width=0.4\linewidth]{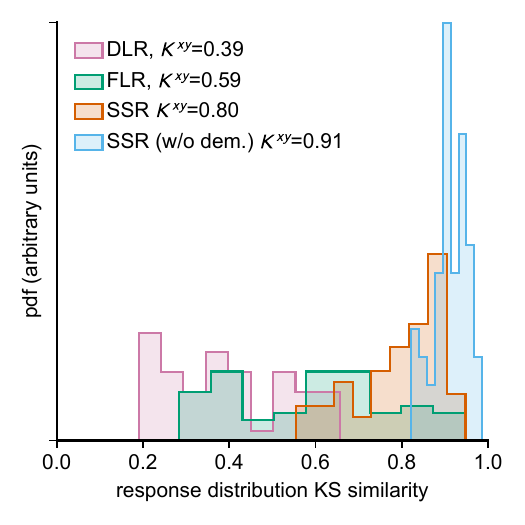} 

    \caption{Comparison of purchase intent distribution similarity between real and synthetic surveys based on \gem{} with $T_\mathrm{LLM}=0.5$ for direct Likert ratings (DLRs), textual elicitation with follow-up Likert ratings (FLRs), semantic similarity ratings (SSRs), and best-set SSRs for an experiment where synthetic consumers where prompted without demographic markers.}
    \label{fig:gem-results-KS-sim}
\end{figure*}

\begin{figure*}
    \centering
    \includegraphics[width=0.45\linewidth]{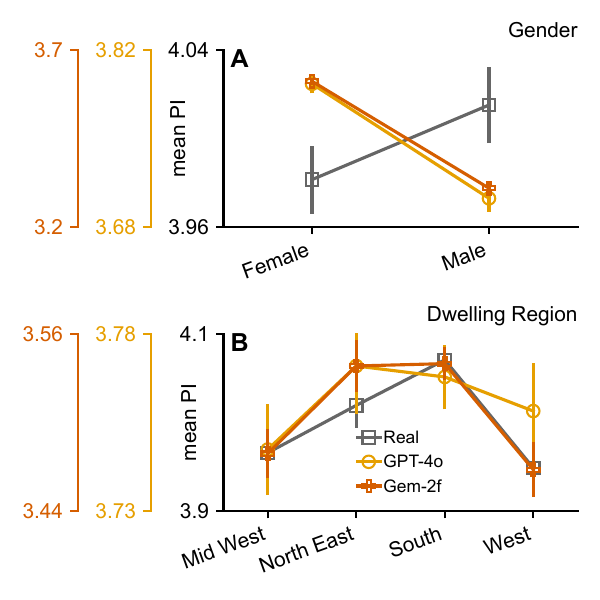} 

    \caption{Mean purchase intent stratified by respondents' gender and dwelling region (shown are results from the SSR method for both \gpt{} and \gem{}). Error bars represent standard errors.}
    \label{fig:gender-region}
\end{figure*}

\begin{figure*}[p] 
    \centering
    \includegraphics[width=1\linewidth]{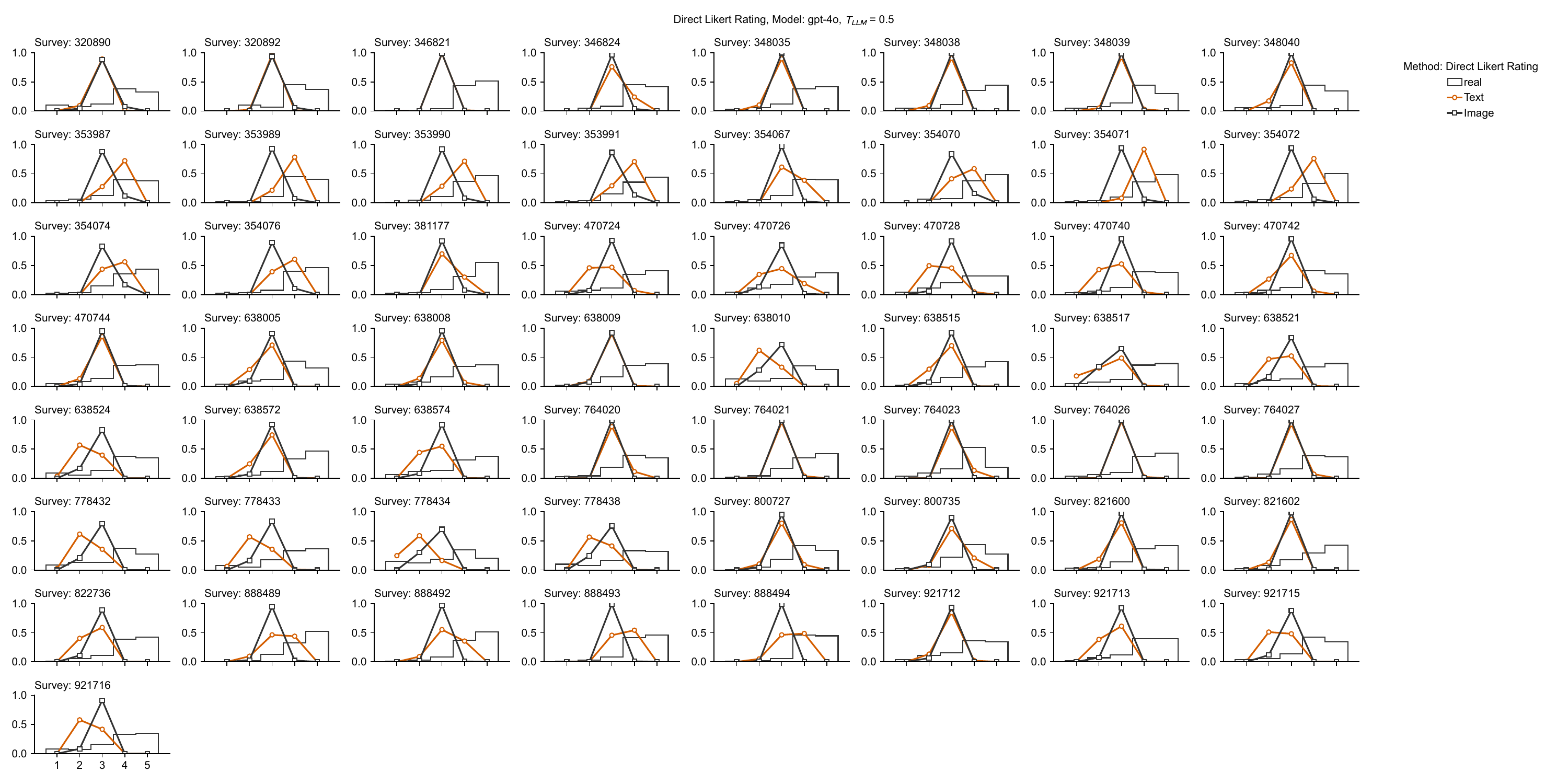} 

    \caption{Survey histograms for direct Likert ratings at $T_\mathrm{LLM}=0.5$ for \gpt{}. }
    \label{fig:DR-hists-gpt-T0.5}
\end{figure*}

\begin{figure*}[p] 
    \centering
    \includegraphics[width=1\linewidth]{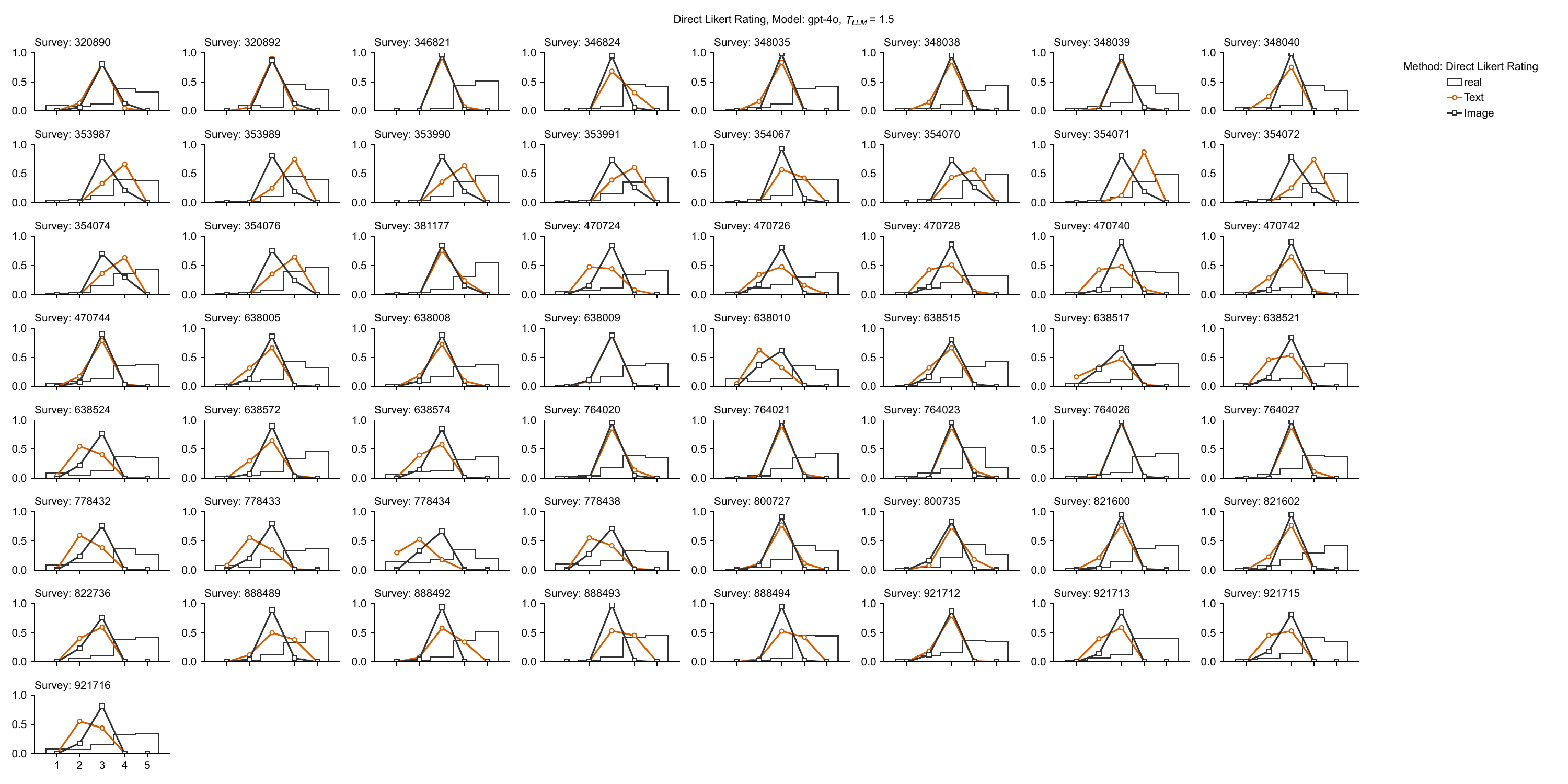} 

    \caption{Survey histograms for direct Likert ratings at $T_\mathrm{LLM}=1.5$ for \gpt{}.}
    \label{fig:DR-hists-gpt-T1.5}
\end{figure*}

\begin{figure*}[p] 
    \centering
    \includegraphics[width=1\linewidth]{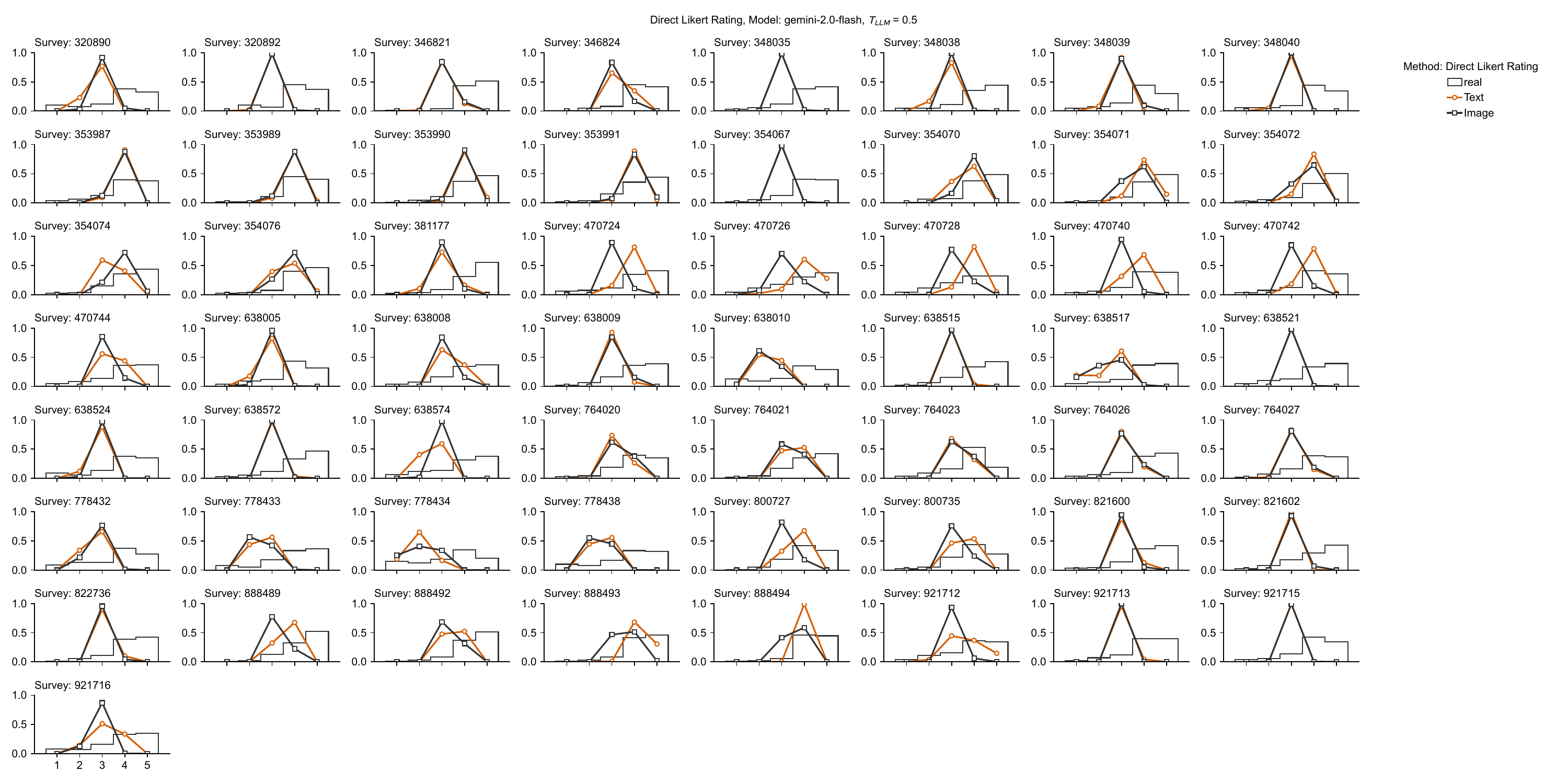} 

    \caption{Survey histograms for direct Likert ratings at $T_\mathrm{LLM}=0.5$ for \gem{}. }
    \label{fig:DR-hists-gem-T0.5}
\end{figure*}

\begin{figure*}[p] 
    \centering
    \includegraphics[width=1\linewidth]{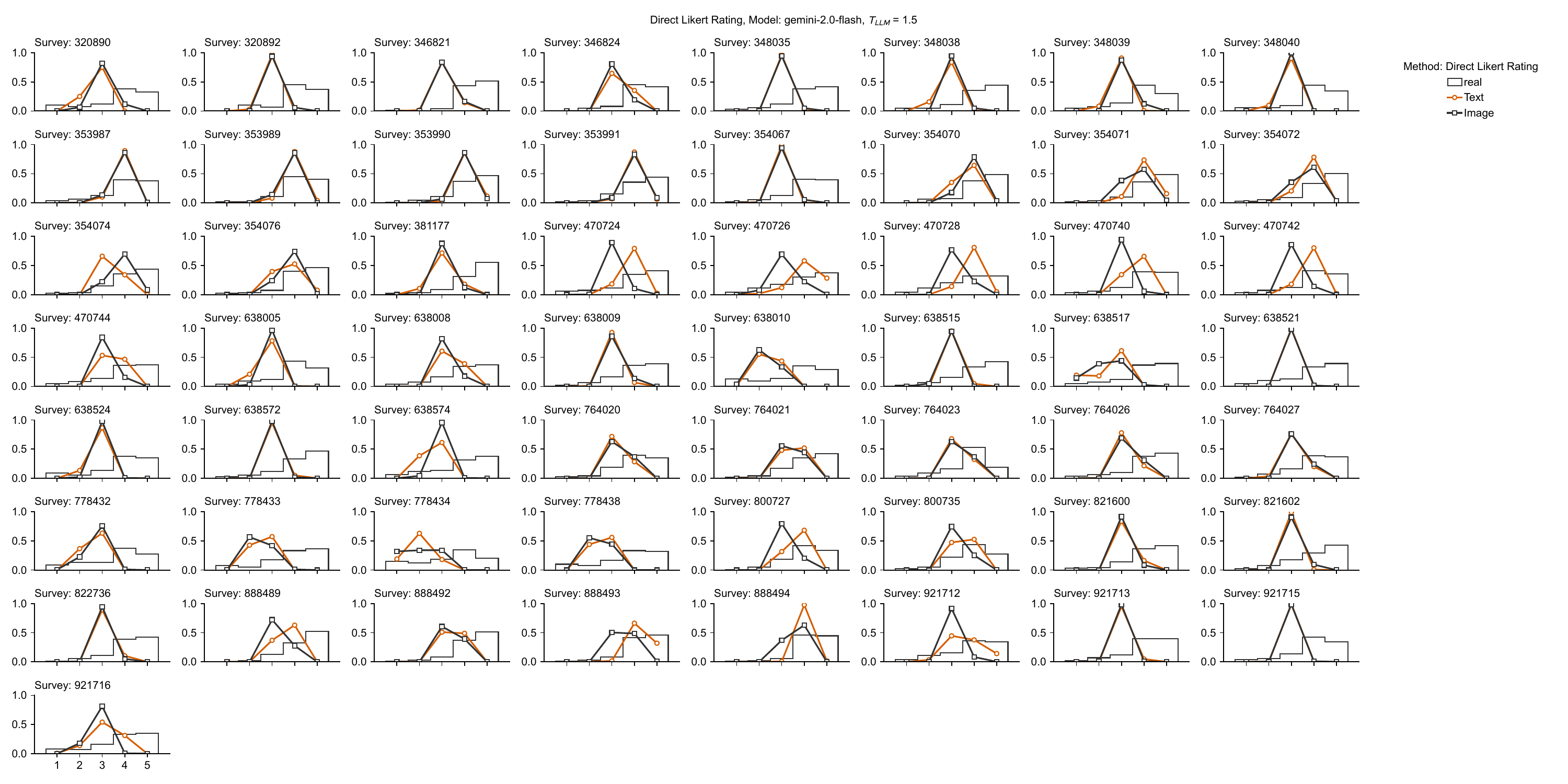} 

    \caption{Survey histograms for direct Likert ratings at $T_\mathrm{LLM}=1.5$ for \gem{}.}
    \label{fig:DR-hists-gem-T1.5}
\end{figure*}

\begin{figure*}
    \centering
    \includegraphics[width=0.5\textwidth]{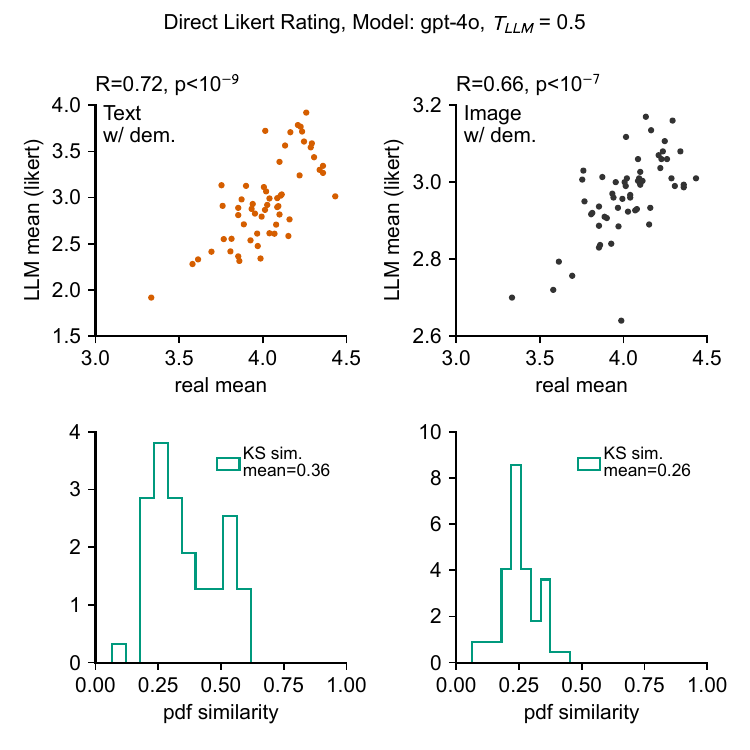} 

    \caption{Success metrics for direct Likert ratings at $T_\mathrm{LLM}=0.5$ for \gpt{}.}
    \label{fig:DR-metrics-gpt-T0.5}
\end{figure*}

\begin{figure*} 
    \centering
    \includegraphics[width=0.5\linewidth]{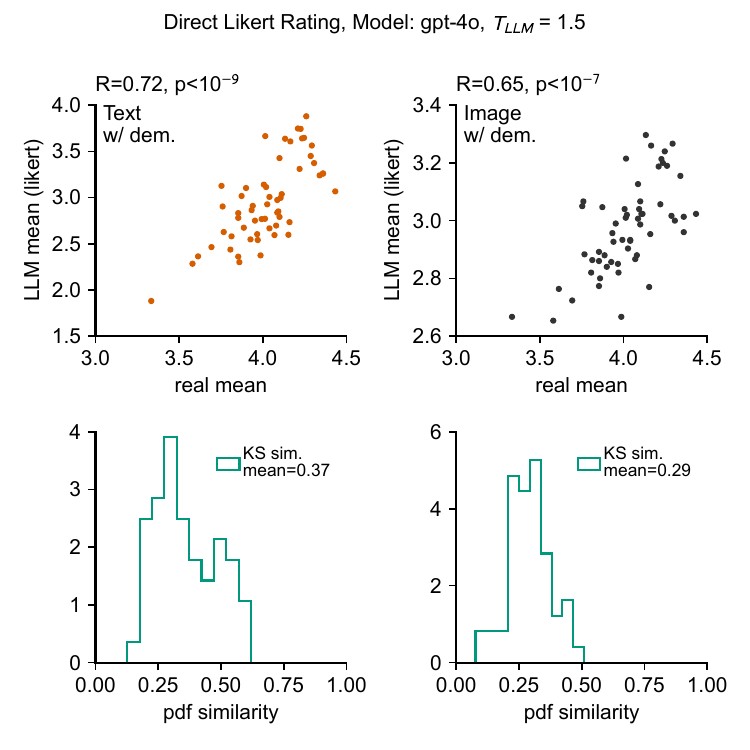} 

    \caption{Success metrics for direct Likert ratings at $T_\mathrm{LLM}=1.5$ for \gpt{}.}
    \label{fig:DR-metrics-gpt-T1.5}
\end{figure*}

\begin{figure*} 
    \centering
    \includegraphics[width=0.5\linewidth]{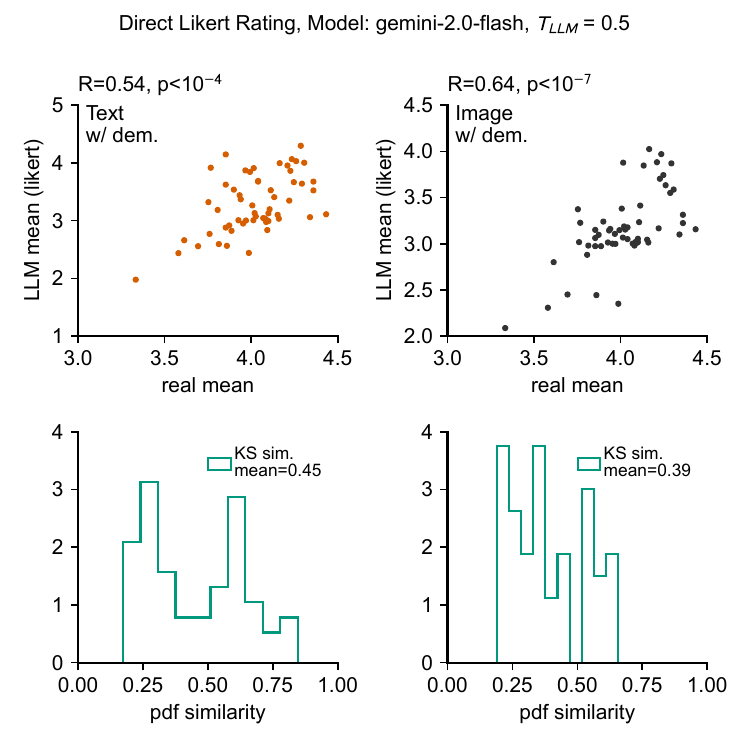} 

    \caption{Success metrics for direct Likert ratings at $T_\mathrm{LLM}=0.5$ for \gem{}. }
    \label{fig:DR-metrics-gem-T0.5}
\end{figure*}

\begin{figure*} 
    \centering
    \includegraphics[width=0.5\linewidth]{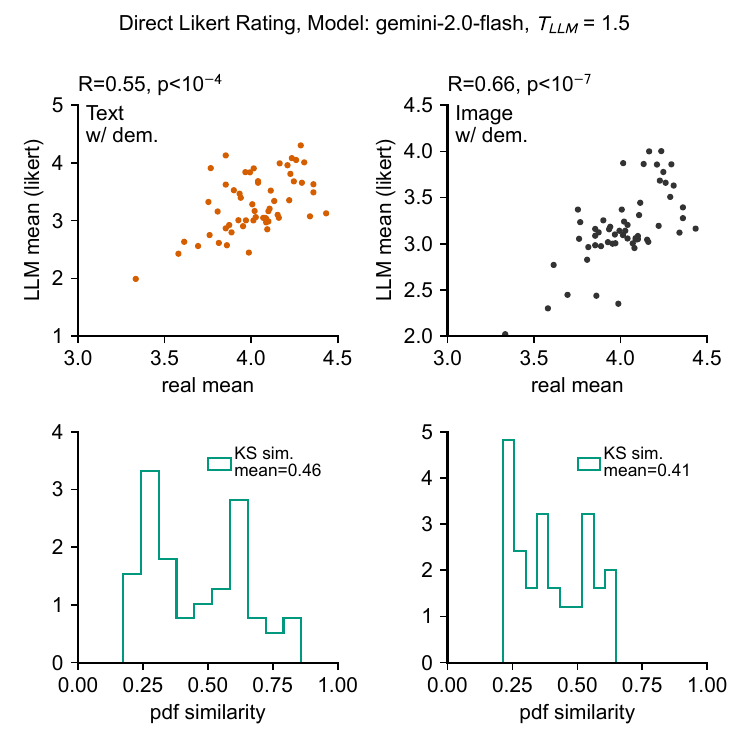} 

    \caption{Success metrics for direct Likert ratings at $T_\mathrm{LLM}=1.5$ for \gem{}.}
    \label{fig:DR-metrics-gem-T1.5}
\end{figure*}

\begin{figure*}[p] 
    \centering
    \includegraphics[width=1\linewidth]{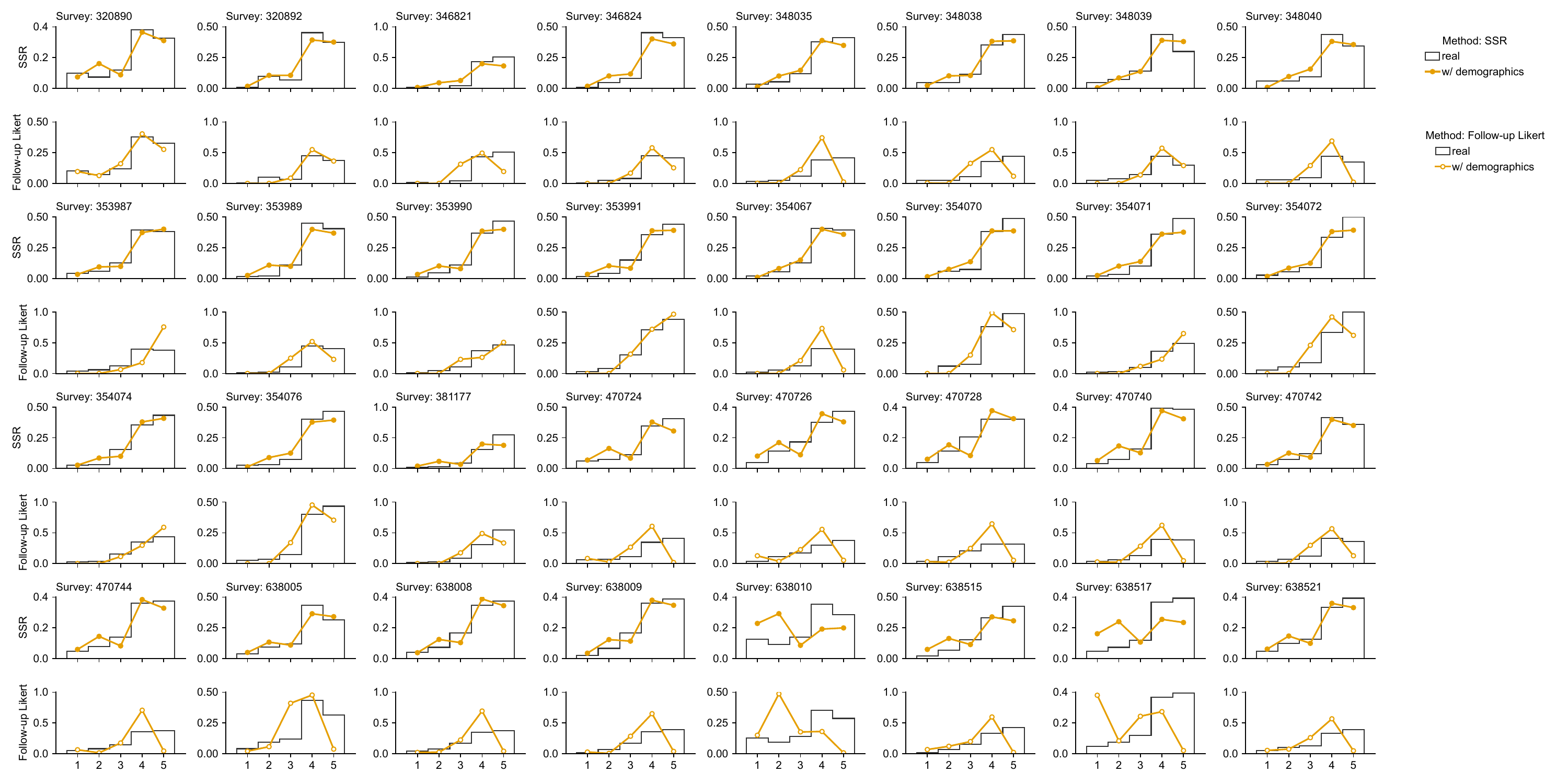} 

    \caption{First set of survey histograms for textual elicitation with \gpt{} and follow-up ratings at $T_\mathrm{LLM}=0.5$, with image stimulus and full demography setup. For semantic similarity rating (SSR), we used the mean over all reference sets.}
    \label{fig:SSR-hists-gpt-T0.5-00}
\end{figure*}

\begin{figure*}[p] 
    \centering
    \includegraphics[width=1\linewidth]{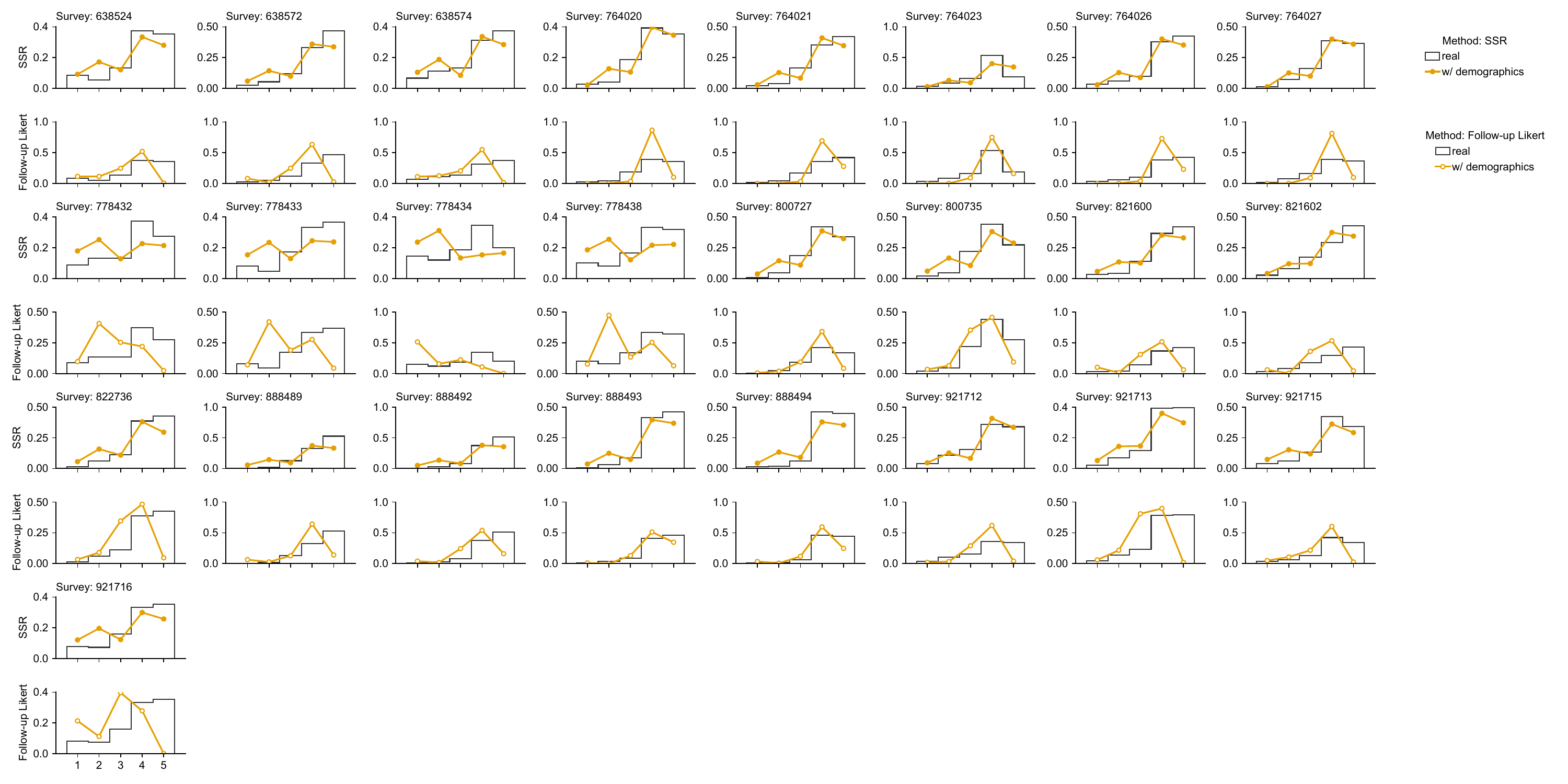} 

    \caption{Second set of survey histograms for textual elicitation with \gpt{} and follow-up ratings at $T_\mathrm{LLM}=0.5$, with image stimulus and full demography setup. For semantic similarity rating (SSR), we used the mean over all reference sets.}
    \label{fig:SSR-hists-gpt-T0.5-01}
\end{figure*}

\begin{figure*}[p] 
    \centering
    \includegraphics[width=1\linewidth]{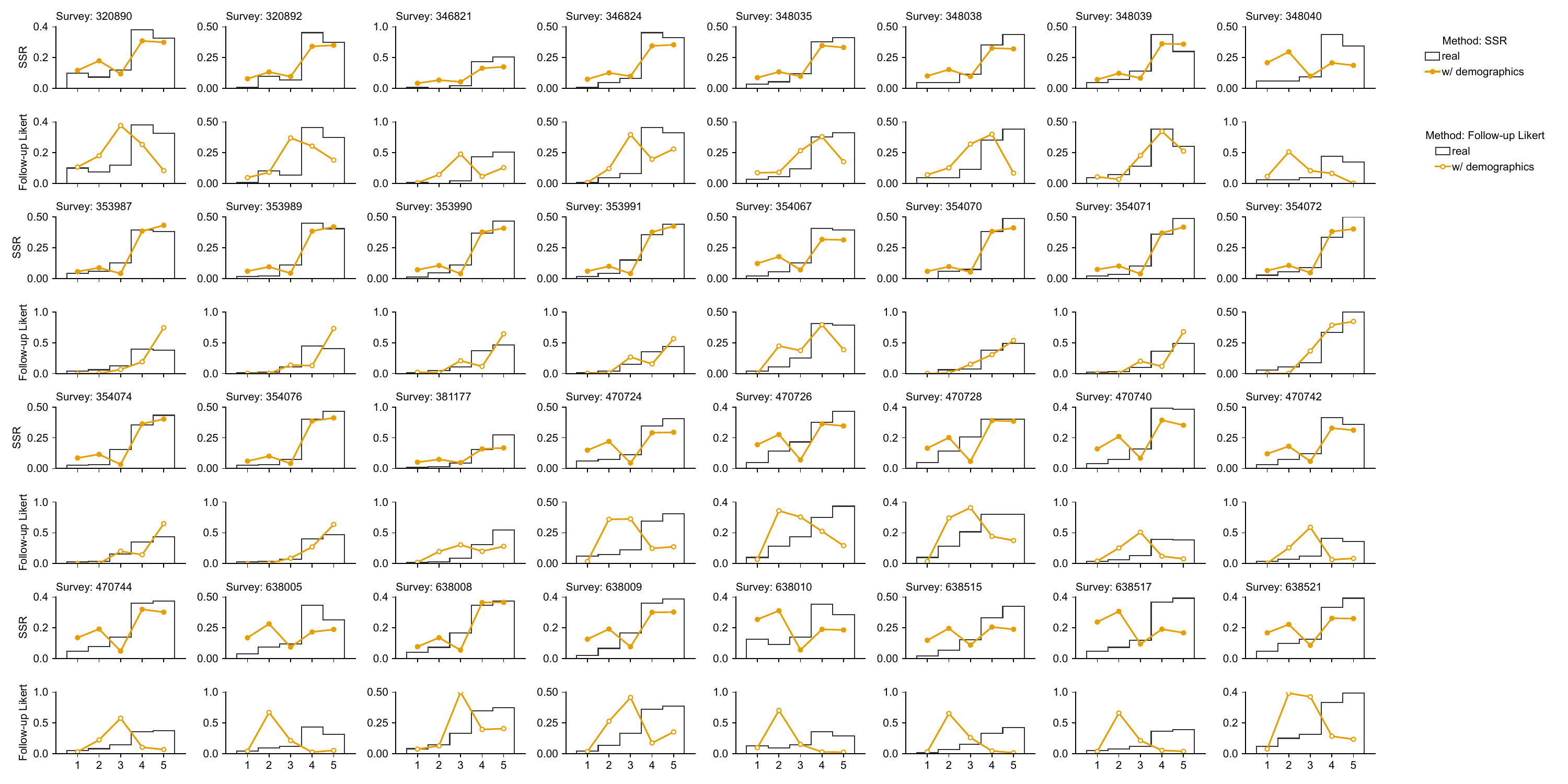} 

    \caption{First set of survey histograms for textual elicitation with \gem{} and follow-up ratings at $T_\mathrm{LLM}=0.5$, with image stimulus and full demography setup. For semantic similarity rating (SSR), we used the mean over all reference sets.}
    \label{fig:SSR-hists-gem-T0.5-00}
\end{figure*}

\begin{figure*}[p] 
    \centering
    \includegraphics[width=1\linewidth]{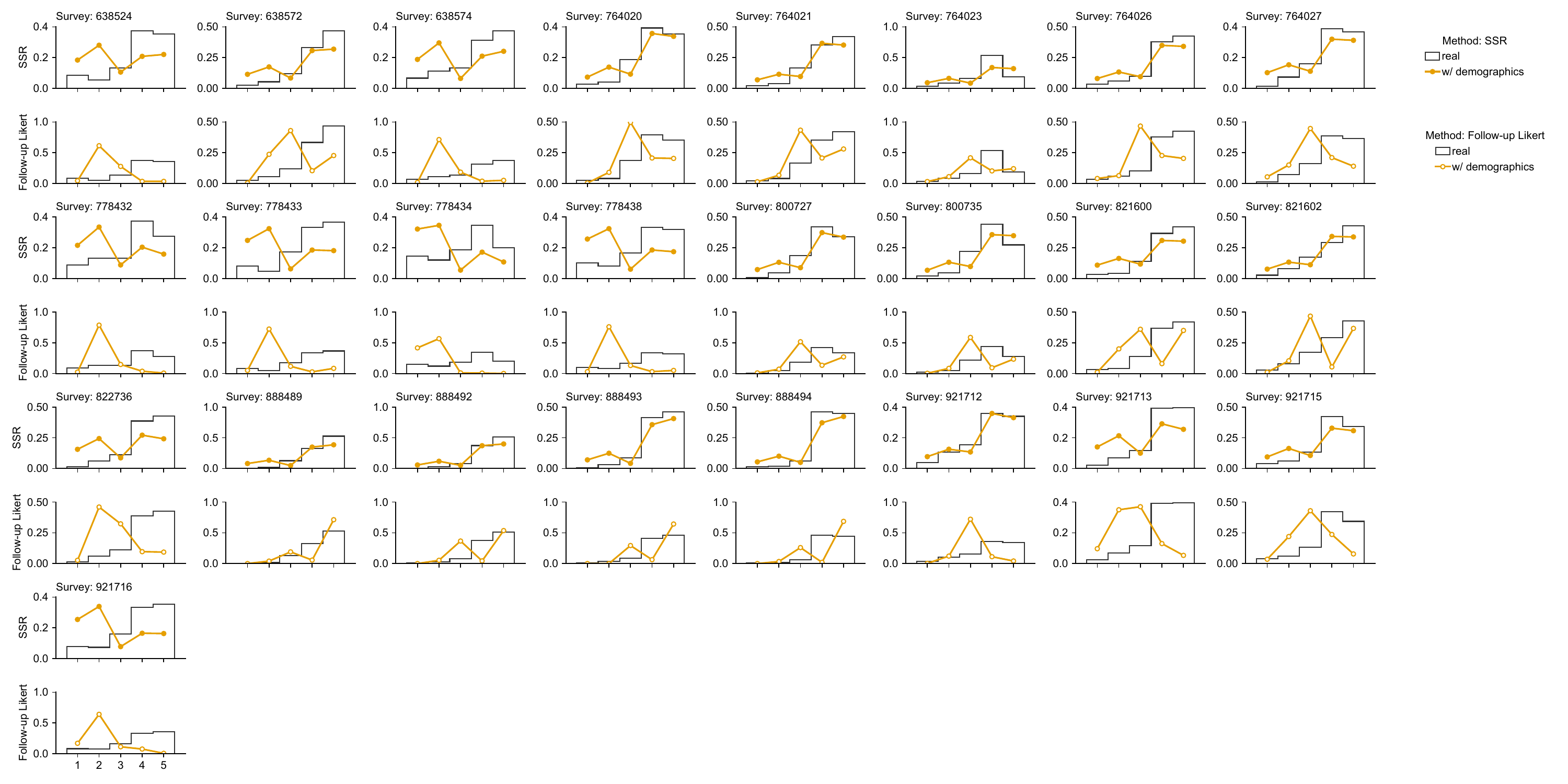} 

    \caption{Second set of survey histograms for textual elicitation with \gem{} and follow-up ratings at $T_\mathrm{LLM}=0.5$, with image stimulus and full demography setup. For semantic similarity rating (SSR), we used the mean over all reference sets.}
    \label{fig:SSR-hists-gem-T0.5-01}
\end{figure*}

\begin{figure*}
    \centering
    \includegraphics[width=0.5\textwidth]{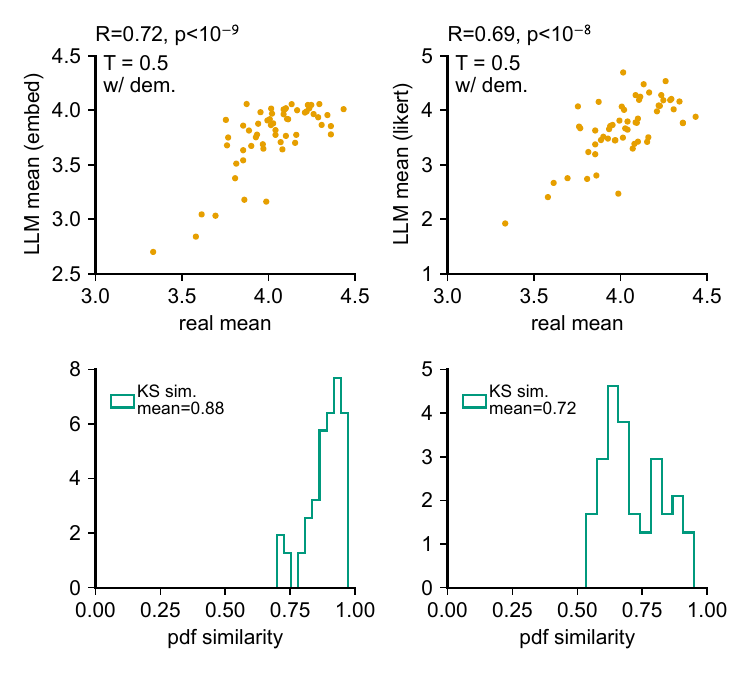} 

    \caption{Success metrics for textual elicitation at $T_\mathrm{LLM}=0.5$ with \gpt{}, with image stimulus and full demography setup. For semantic similarity rating (SSR), we used the mean over all reference sets.}
    \label{fig:SSR-metrics-gpt-T0.5}
\end{figure*}

\begin{figure*}[p] 
    \centering
    \includegraphics[width=0.5\linewidth]{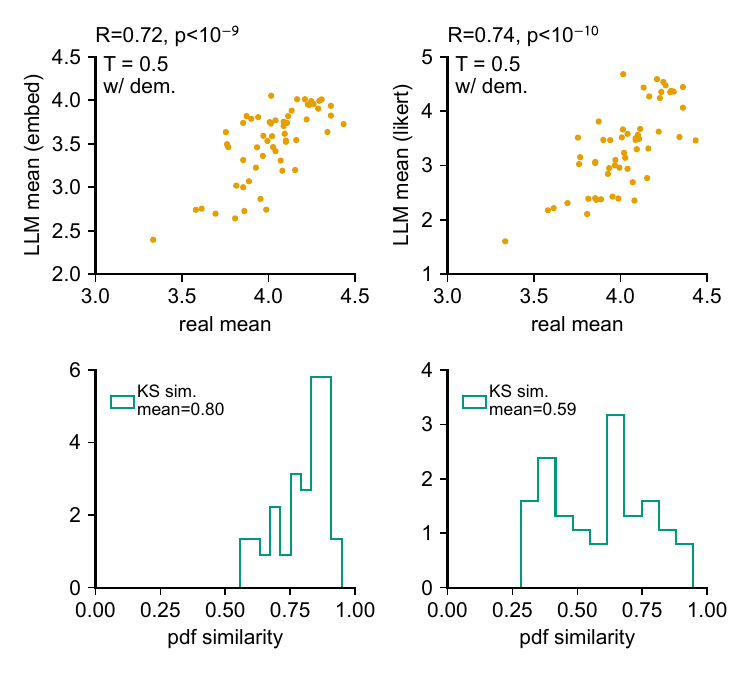} 

    \caption{Success metrics for textual elicitation at $T_\mathrm{LLM}=0.5$ with \gem{}, with image stimulus and full demography setup. For semantic similarity rating (SSR), we used the mean over all reference sets.}
    \label{fig:SSR-metrics-gem-T0.5}
\end{figure*}

\begin{figure*}[p] 
    \centering
    \includegraphics[width=1\linewidth]{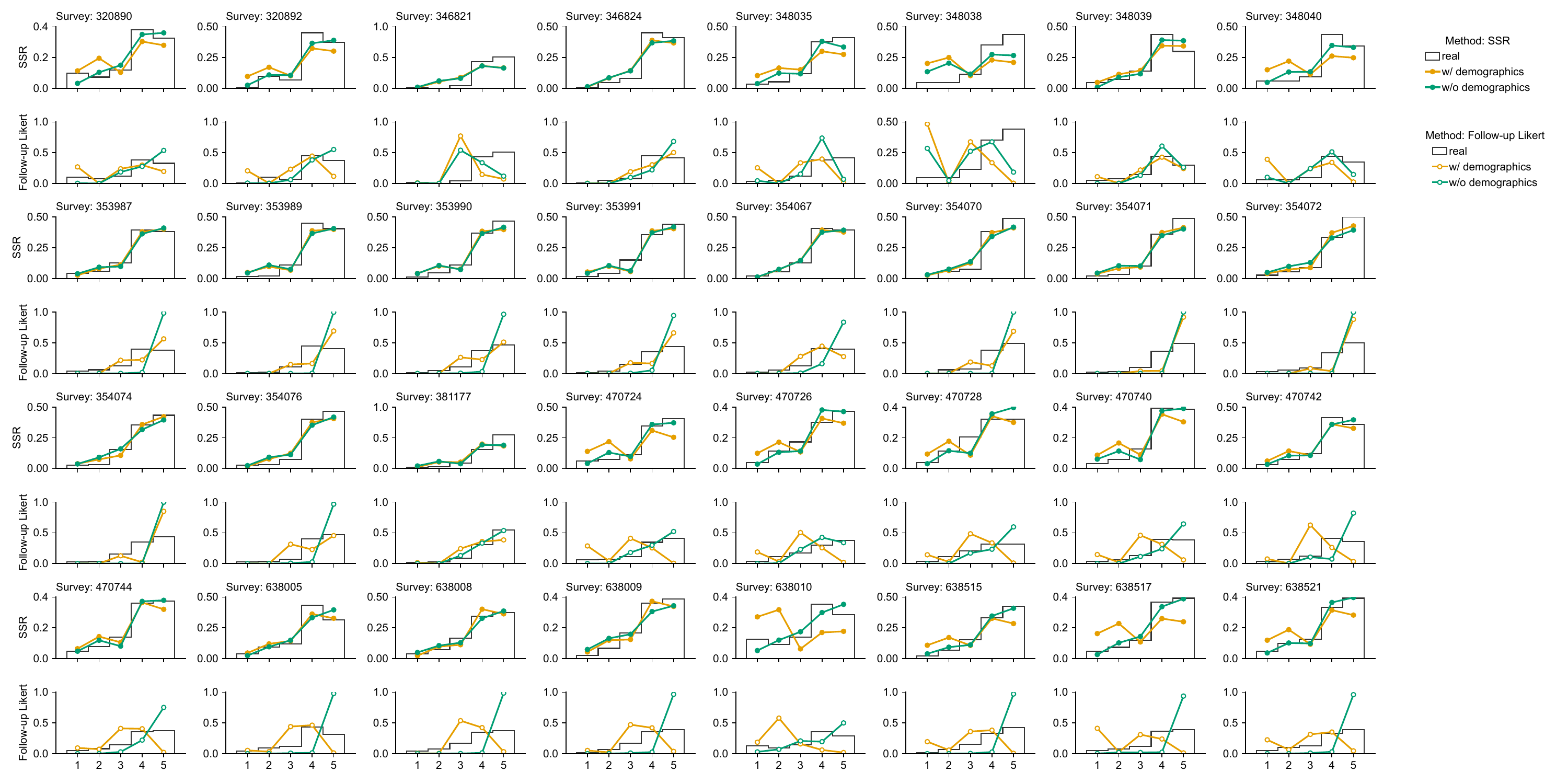} 

    \caption{First set of survey histograms for textual elicitation with \gpt{} and follow-up ratings at $T_\mathrm{LLM}=0.5$, with text stimulus and alternating between prompting the LLM with \emph{full} demographic information and \emph{zero} demographic information. For semantic similarity rating (SSR), we used the mean over all reference sets.}
    \label{fig:demog-hists-gpt-T0.5-00}
\end{figure*}

\begin{figure*}[p] 
    \centering
    \includegraphics[width=1\linewidth]{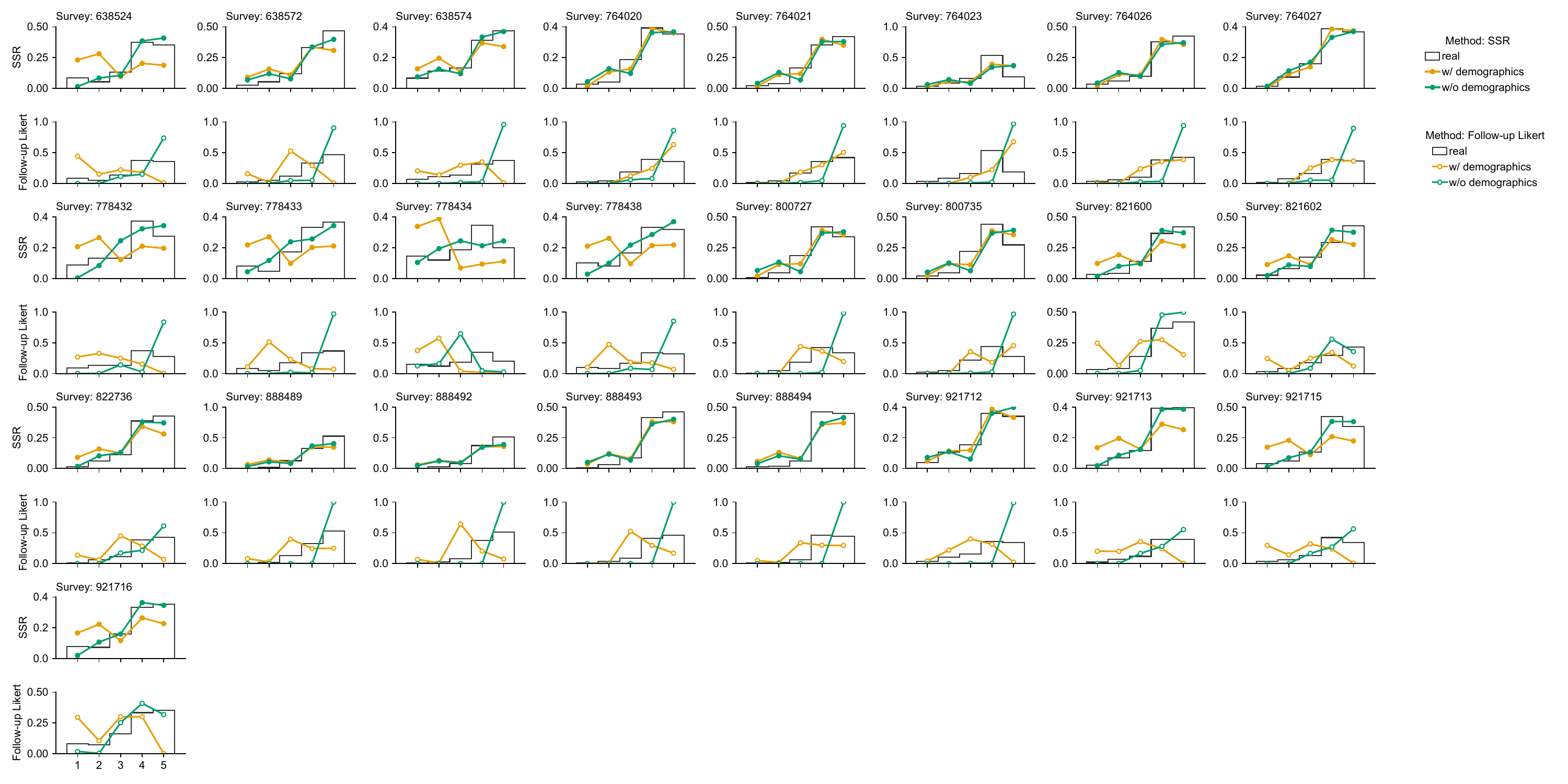} 

    \caption{Second set of survey histograms for textual elicitation with \gpt{} and follow-up ratings at $T_\mathrm{LLM}=0.5$, with text stimulus and alternating between prompting the LLM with \emph{full} demographic information and \emph{zero} demographic information. For semantic similarity rating (SSR), we used the mean over all reference sets.}
    \label{fig:demog-hists-gpt-T0.5-01}
\end{figure*}

\begin{figure*}[p] 
    \centering
    \includegraphics[width=1\linewidth]{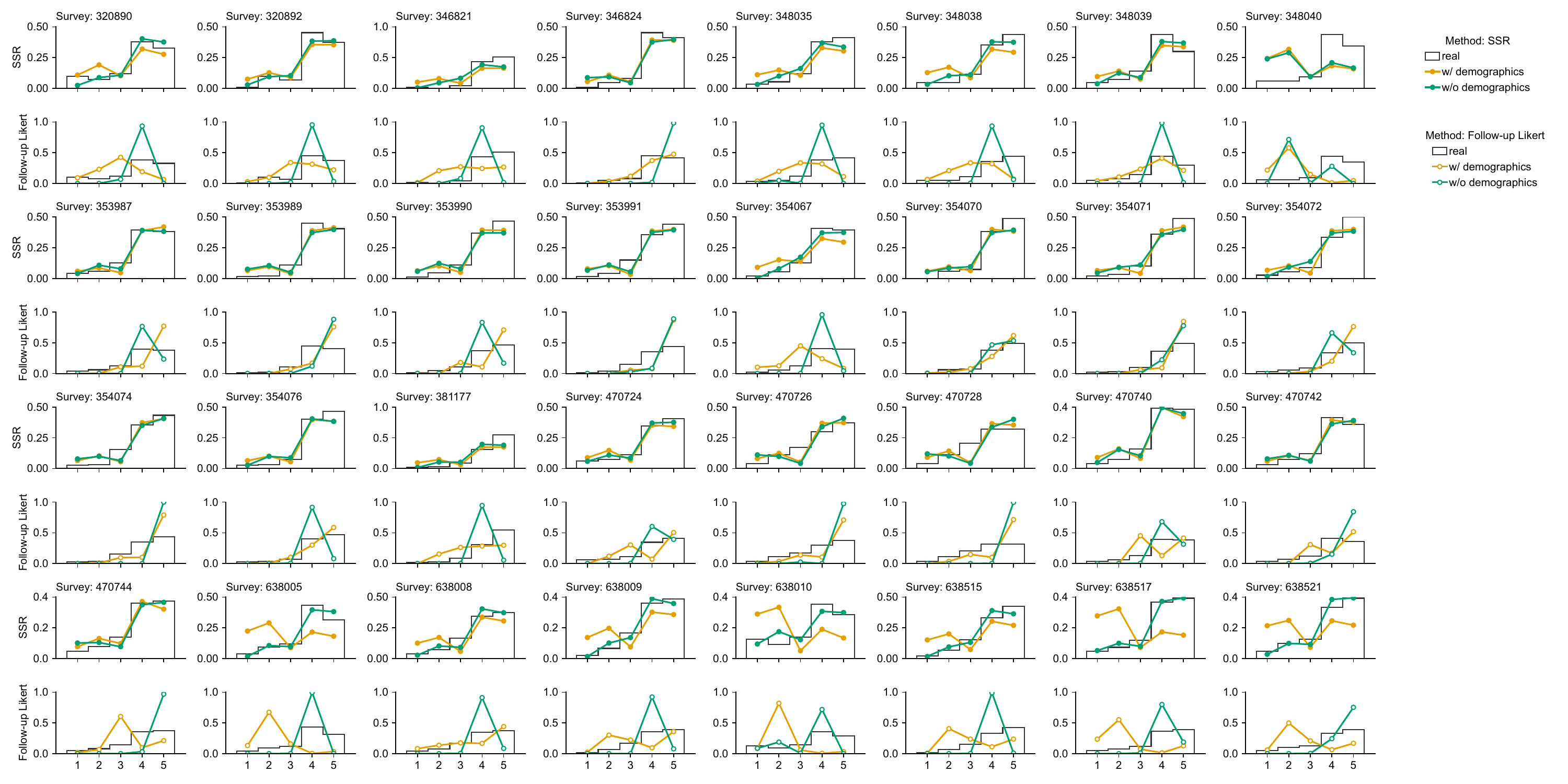} 

    \caption{First set of survey histograms for textual elicitation with \gem{} and follow-up ratings at $T_\mathrm{LLM}=0.5$, with text stimulus and alternating between prompting the LLM with \emph{full} demographic information and \emph{zero} demographic information. For semantic similarity rating (SSR), we used the mean over all reference sets.}
    \label{fig:demog-hists-gem-T0.5-00}
\end{figure*}

\begin{figure*}[p] 
    \centering
    \includegraphics[width=1\linewidth]{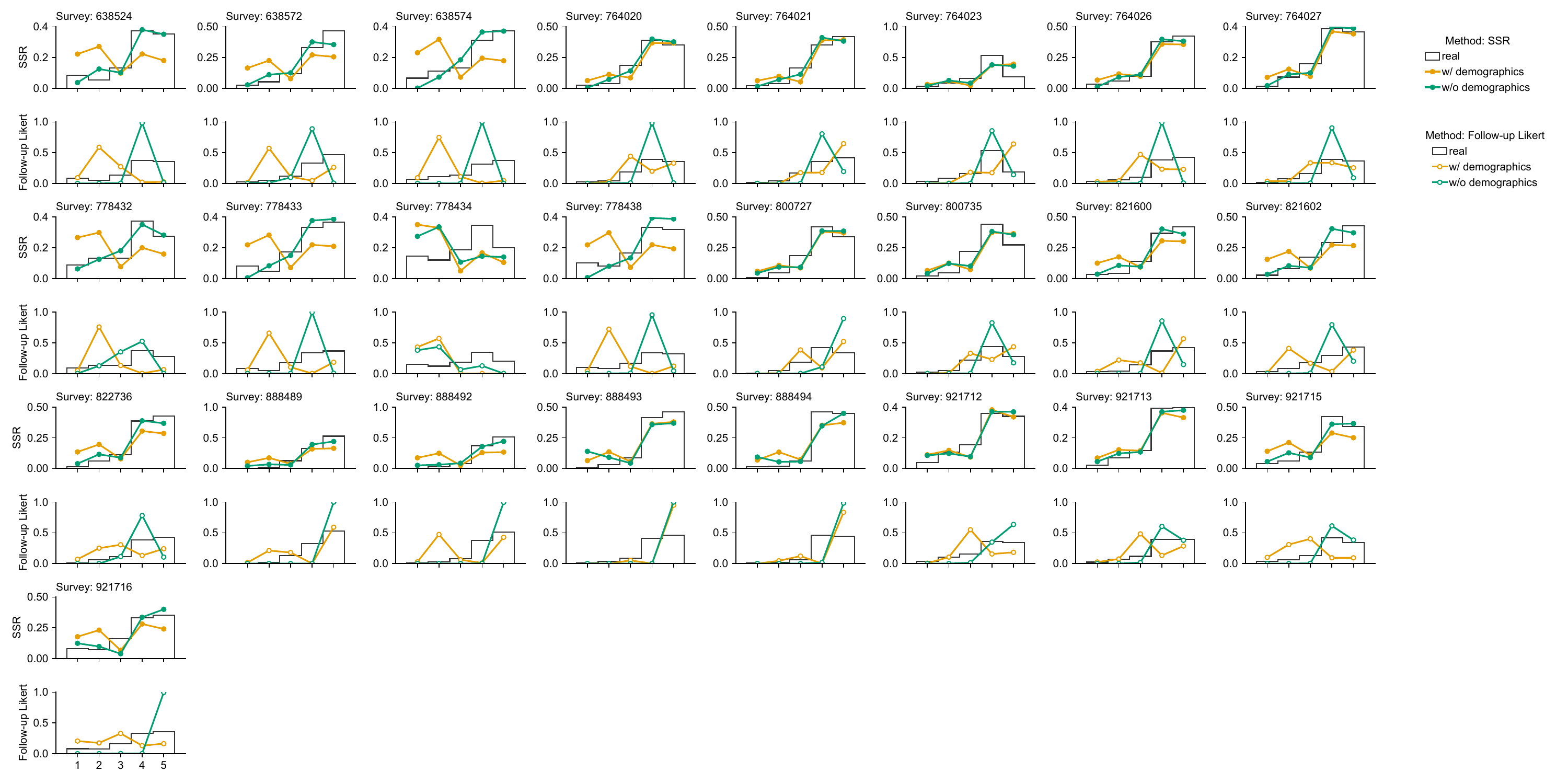} 

    \caption{Second set of survey histograms for textual elicitation with \gem{} and follow-up ratings at $T_\mathrm{LLM}=0.5$, with text stimulus and alternating between prompting the LLM with \emph{full} demographic information and \emph{zero} demographic information. For semantic similarity rating (SSR), we used the mean over all reference sets.}
    \label{fig:demog-hists-gem-T0.5-01}
\end{figure*}

\begin{figure*}
    \centering
    \includegraphics[width=1\textwidth]{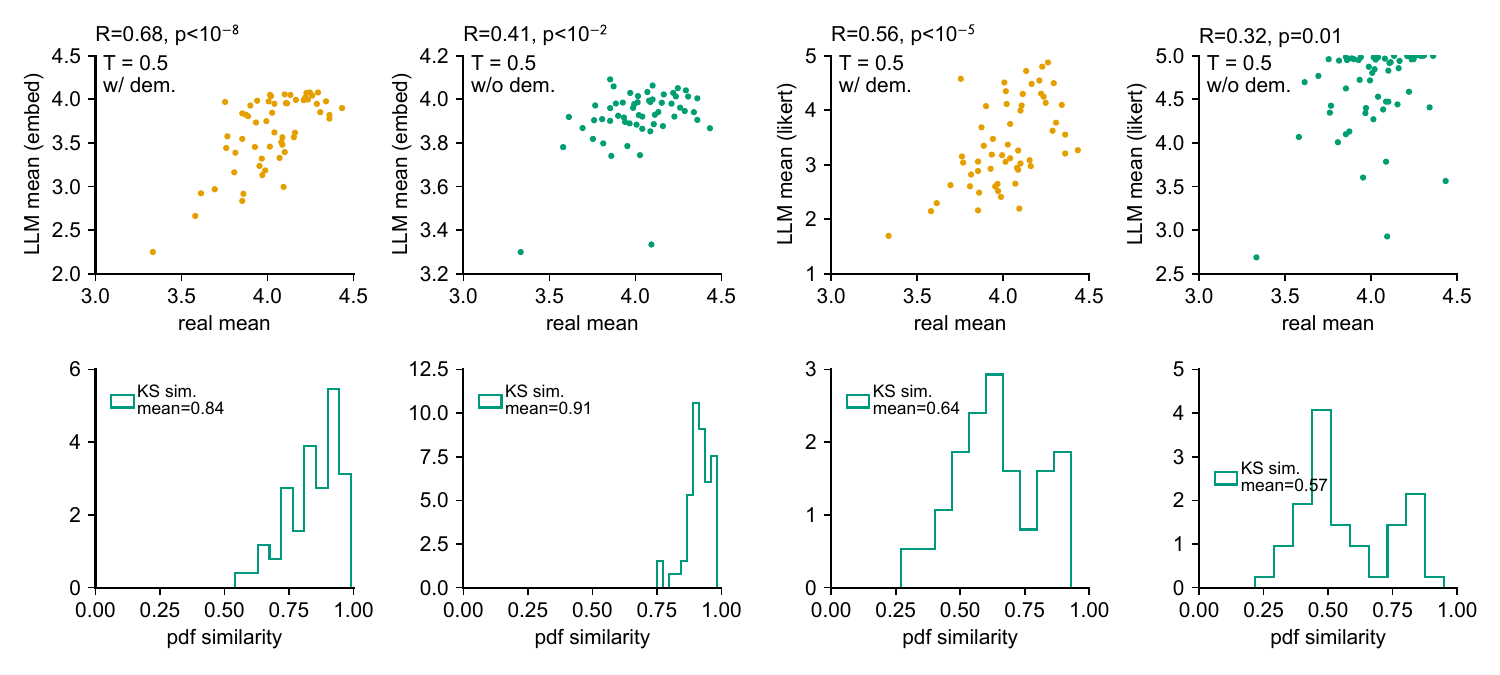} 

    \caption{Success metrics for textual elicitation and demography experiments, at $T_\mathrm{LLM}=0.5$ with \gpt{} and with text stimulus. For semantic similarity rating (SSR), we used the mean over all reference sets.}
    \label{fig:demog-metrics-gpt-T0.5}
\end{figure*}

\begin{figure*}
    \centering
    \includegraphics[width=1\textwidth]{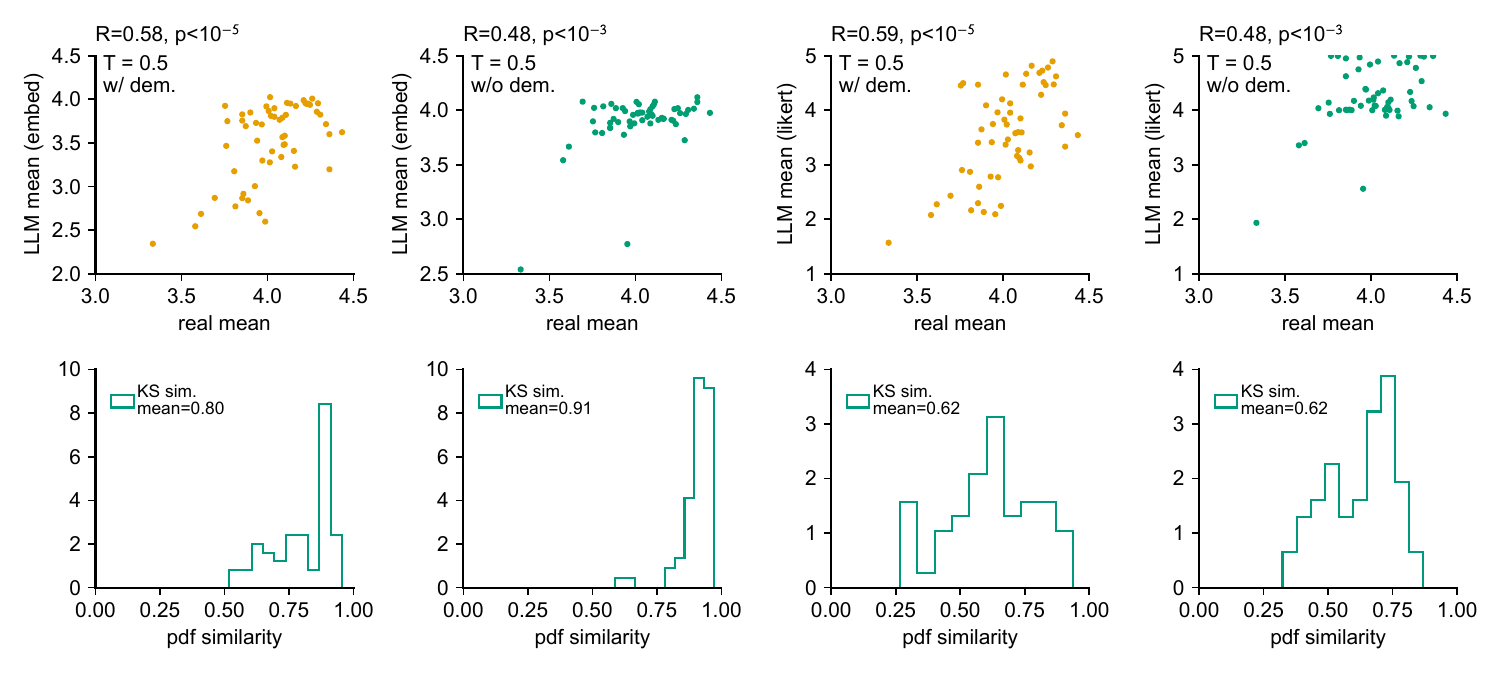} 

    \caption{Success metrics for textual elicitation and demography experiments, at $T_\mathrm{LLM}=0.5$ with \gem{} and with text stimulus. For semantic similarity rating (SSR), we used the mean over all reference sets.}
    \label{fig:demog-metrics-gem-T0.5}
\end{figure*}
\begin{figure*}[p]
    \centering
    \includegraphics[width=0.5\textwidth]{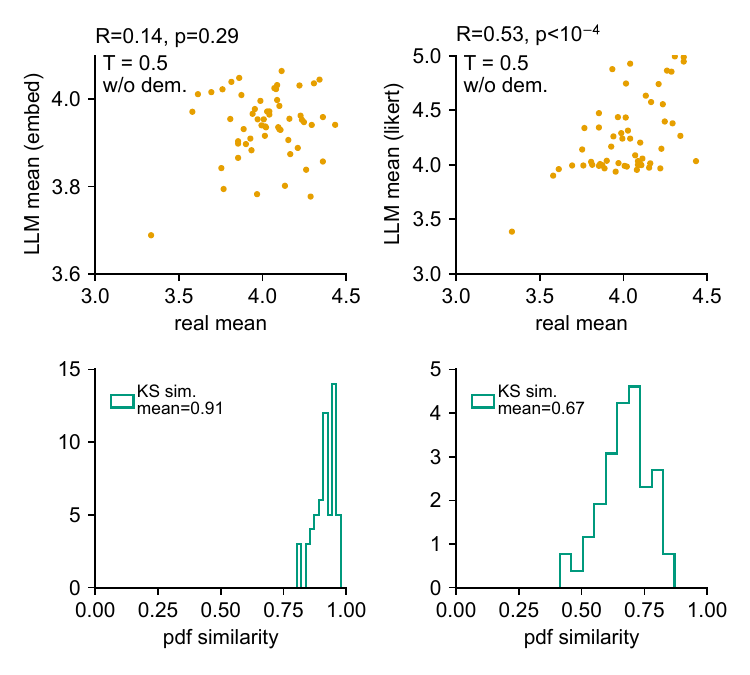} 

    \caption{Success metrics for textual elicitation and demography experiments, at $T_\mathrm{LLM}=0.5$ with \gem{} and image stimulus. For semantic similarity rating (SSR), we used the result for the best reference set (4).}
    \label{fig:wodemog-image-metrics-gem-T0.5-00} 
\end{figure*}

\begin{figure*}[p] 
    \centering
    \includegraphics[width=1\linewidth]{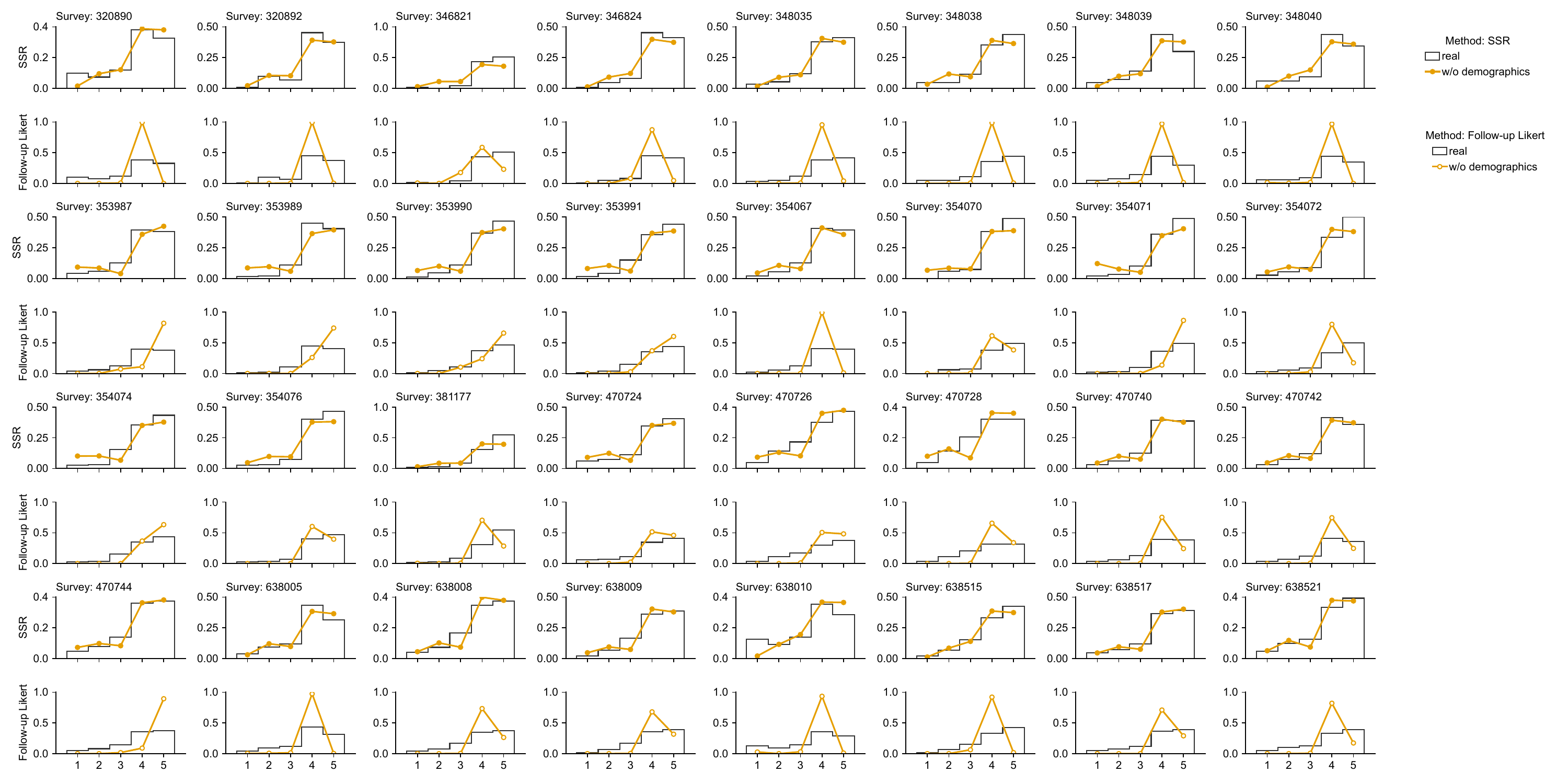} 

    \caption{First set of survey histograms for textual elicitation with \gem{} and follow-up ratings at $T_\mathrm{LLM}=0.5$, with image stimulus and prompting the LLM with \emph{zero} demographic information. For semantic similarity rating (SSR), we used the the result for the best reference set (4).}
    \label{fig:wodemog-image-hists-gem-T0.5-00}
\end{figure*}

\begin{figure*}[p] 
    \centering
    \includegraphics[width=1\linewidth]{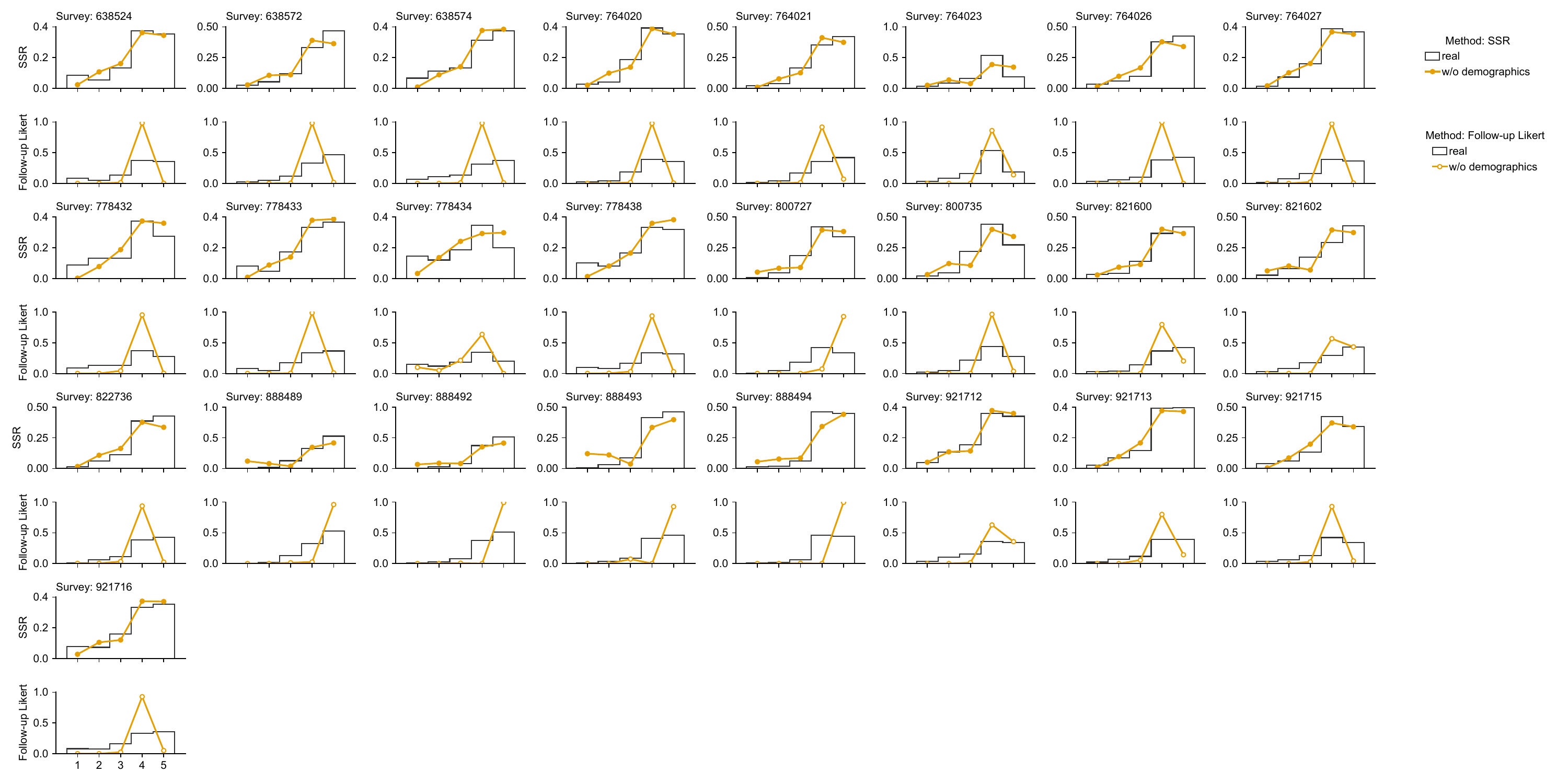} 

    \caption{Second set of survey histograms for textual elicitation with \gem{} and follow-up ratings at $T_\mathrm{LLM}=0.5$, with image stimulus and prompting the LLM with \emph{zero} demographic information. For semantic similarity rating (SSR), we used the the result for the best reference set (4).}
    \label{fig:wodemog-image-hists-gem-T0.5-01}
\end{figure*}

\begin{figure*}[p]
    \centering
    \includegraphics[width=1\linewidth]{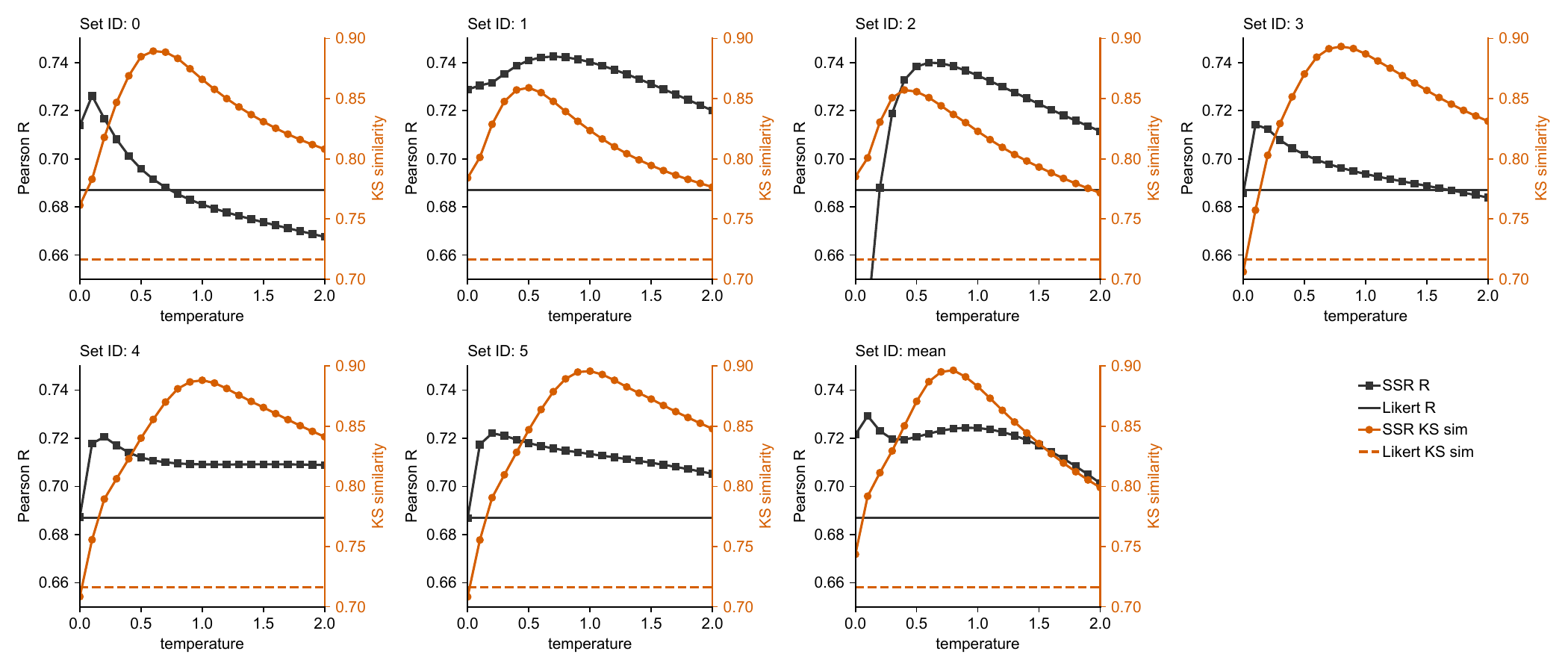} 

    \caption{Scan over post-elicitation temperature $T$ values and change in success metrics for textual elicitation at $T_\mathrm{LLM}=0.5$ with \gpt{} and image stimulus, with full demography setup. For semantic similarity rating (SSR), we used the mean over all reference sets. Horizontal lines refer to the corresponding follow-up Likert rating values for this experiment, which are unaffected by choice of reference set and $T$.}
    \label{fig:temperature-scan-SSR-gpt-T0.5}
\end{figure*}
\begin{figure*}[p] 
    \centering
    \includegraphics[width=1\linewidth]{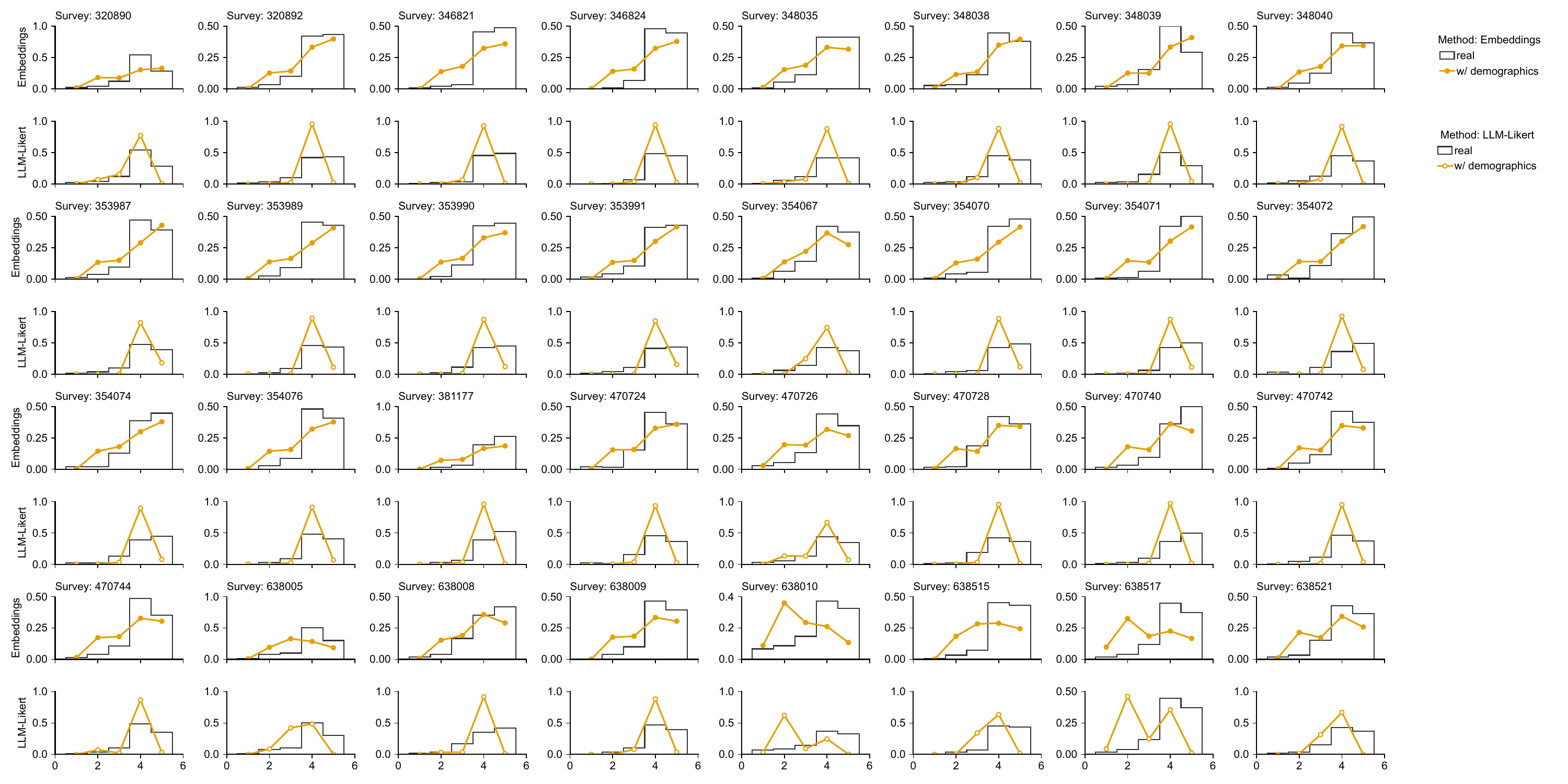} 

    \caption{First set of survey histograms for textual elicitation to question ``How relevant is this concept for you?'' with \gem{} and follow-up ratings at $T_\mathrm{LLM}=0.5$, with image stimulus and full demography setup. For semantic similarity rating (SSR), we used the mean over three reference sets that were constructed for this question specifically. }
    \label{fig:relevance-hists-gem-T0.5-00}
\end{figure*}

\begin{figure*}[p] 
    \centering
    \includegraphics[width=1\linewidth]{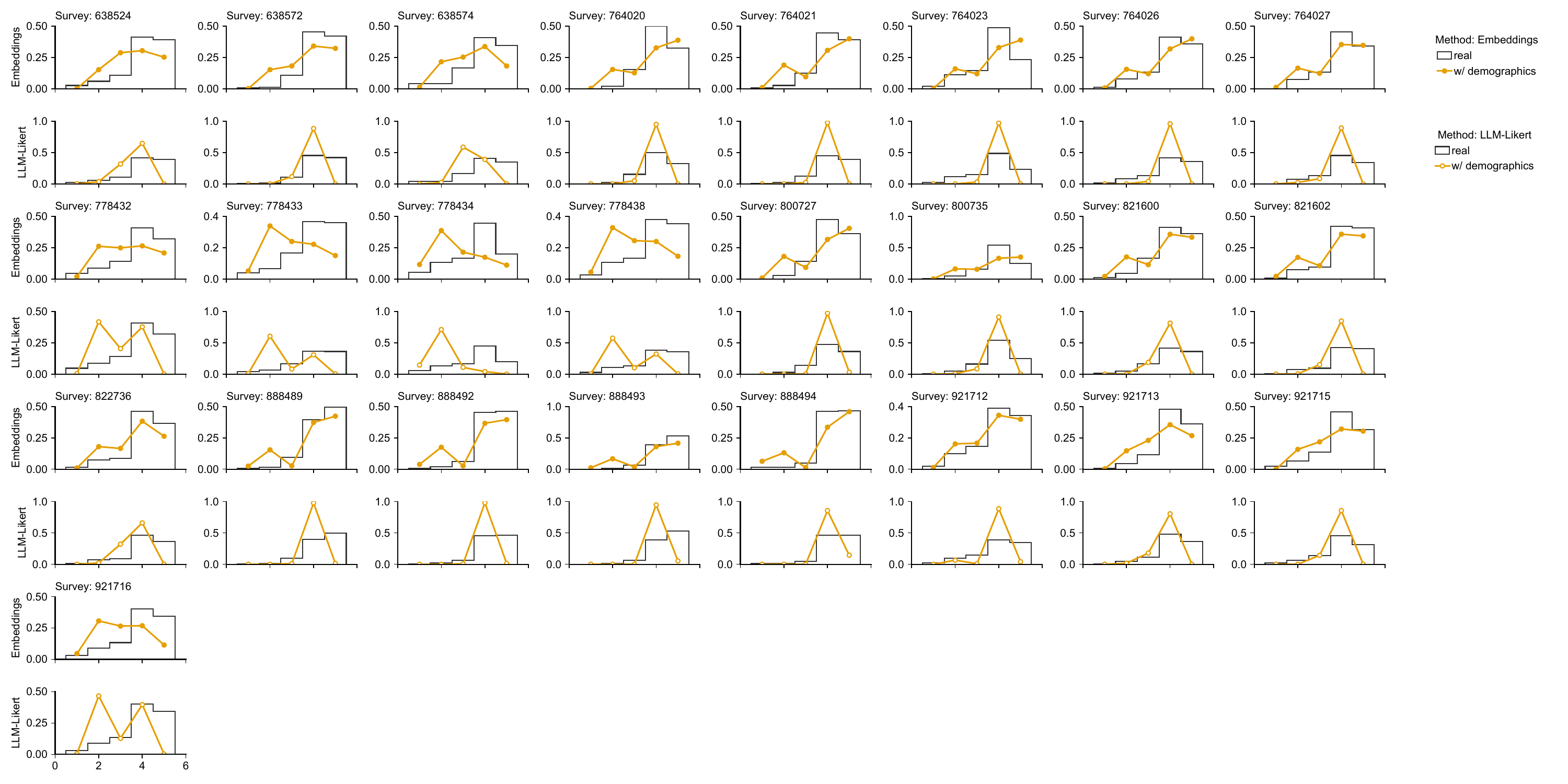} 

    \caption{Second set of survey histograms for textual elicitation to question ``How relevant is this concept for you?'' with \gem{} and follow-up ratings at $T_\mathrm{LLM}=0.5$, with image stimulus and full demography setup. For semantic similarity rating (SSR), we used the mean over three reference sets that were constructed for this question specifically. }
        \label{fig:relevance-hists-gem-T0.5-01}
\end{figure*}

\begin{figure*}
    \centering
    \includegraphics[width=0.5\textwidth]{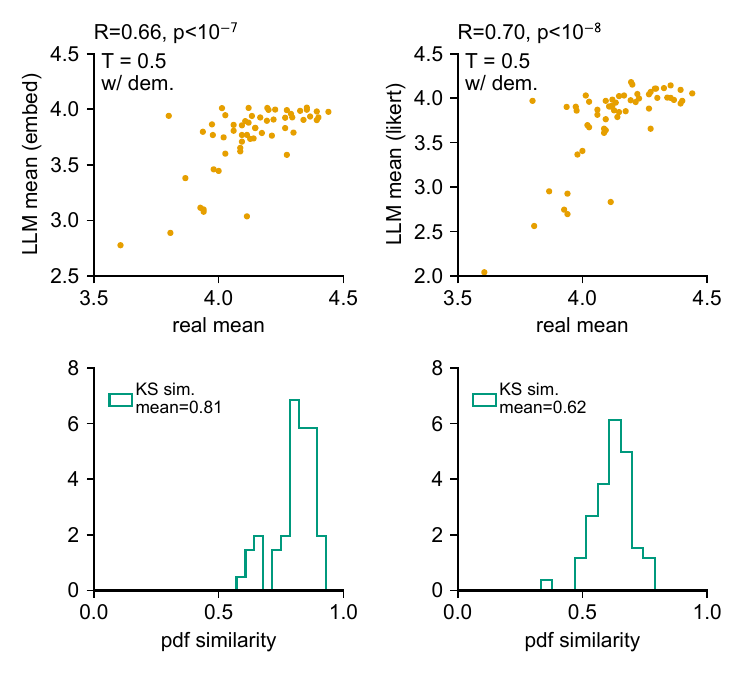} 

    \caption{Success metrics for textual elicitation to question ``How relevant is this concept for you?'' with \gem{} and follow-up ratings at $T_\mathrm{LLM}=0.5$, with image stimulus and full demography setup. For semantic similarity rating (SSR), we used the mean over three reference sets that were constructed for this question specifically.}
    \label{fig:relevance-metrics-gem-T0.5}
\end{figure*}

\end{document}